\documentclass[final,5p,times,twocolumn,authoryear,nopreprintline]{elsarticle} 

\usepackage{amssymb}
\usepackage{amsmath}

\usepackage{subfigure}
\usepackage{url} 
\usepackage{xcolor} 
\usepackage{threeparttable}
\usepackage{placeins}
\usepackage{adjustbox}
\usepackage{float}
\usepackage{manyfoot} 
\usepackage{booktabs} 
\usepackage{multirow} %
\usepackage{makecell}  
\usepackage[shortlabels]{enumitem} 
\usepackage[hidelinks]{hyperref}
\usepackage{orcidlink}

\begin{document}

\begin{frontmatter}

\title{Cropland PAtteRNS: Parallel Dimensional Attention Networks and Attention to Dataset Disparity for Crop Segmentation in Satellite Imagery Time Series Data} 

\author[SU]{Joseph Metcalfe\fnref{Corresponding Author}\orcidlink{0009-0007-4839-5549}} 
\ead{j.metcalfe.2275333@swansea.ac.uk}
\author[SU]{Sara Sharifzadeh\orcidlink{0000-0003-4621-2917}} 
\ead{Sara.Sharifzadeh@swansea.ac.uk}
\author[SU]{Fabio Caraffini\orcidlink{0000-0001-9199-7368}} 
\ead{Fabio.Caraffini@swansea.ac.uk}

\affiliation[SU]{organization={Department of Computer Science, Swansea University},
            city={Swansea},
            postcode={SA1 8EN},
            country={United Kingdom}}

\fntext[corresponding]{Corresponding Author}

\begin{abstract}

The landscape of satellite imagery time series datasets and boundary-pushing architectures for cropland segmentation has never been richer. However, in this gold rush, important truths are being missed on both fronts, as a drive for the most novel concepts or the largest datasets pushes finer details to the side. In this paper, we present our hybrid transformer-convolutional model, Cropland \textbf{P}arallel \textbf{Atte}ntion and \textbf{R}efinement \textbf{N}etwork for \textbf{S}egmentation (Cropland PAtteRNS), the first model to use self-attention mechanisms separately for each of the temporal, spectral, and spatial aspects of Sentinel-2 multispectral SITS data. To achieve fully-factorised attention in our proposed model, we introduce a novel parallel transformer architecture which significantly reduces the computational complexity of triple-factorised self-attention. We validate our architecture with an in-depth ablation study, and analyse the performance of our model against state-of-the-art crop segmentation models on multiple tile-size variants of the popular PASTIS and MTLCC datasets. Our findings show our model to outperform all others in the task of crop class segmentation, verified across multiple important segmentation metrics, with especially strong performance against compared models seen in the often under-reported parcel delineation quality, for which we use the Boundary IoU metric. We also find that flawed class groupings within datasets can have a significant negative impact on model performance, and report that alternate tile-size variants of crop segmentation datasets produce results incomparable to one-another, invalidating fair comparison between model performance when trained on different tile-sizes. Based on these findings, we suggest further work is required to standardise best practices for multiple aspects of constructing SITS crop segmentation datasets, and to identify any relations enabling future dynamic-tile-sizing for ideal model performance.

\end{abstract}

\newpageafter{abstract}


\end{frontmatter}

\section{Introduction}\label{sec:intro}

Crop type mapping is a long-established field in earth observation (EO) remote sensing, with essential and wide-ranging applications. From the first dedicated EO satellite mission, Landsat 1, a key advantage of the spawned Satellite Imagery Time Series (SITS) data products has been their potential to increase the dimensionality of available imagery data through multispectral and multi-temporal captures \citep{wulderLandsatRev}. This lends SITS naturally to agricultural applications, where growth-cycle variations can only be captured within  the temporal dimension, and the increasingly well-characterised discriminative power of banded spectral absorbance differences between plant species \citep{thenkabail2012hyperspectral}. Modern advances in SITS data quality and accessibility in both commercial and public domains have enabled numerous studies across agricultural, environmental, surveillance, and many other fields. These developments have paralleled a surge in modern machine learning (ML) capabilities. As data landscapes outgrew feasible widespread manual classification, the use of ML in remote sensing has rapidly grown, with applications generally falling into the three categories of 1) \textit{Classification}: assigning a single image-level class to a whole tile or region of interest (ROI); 2) \textit{Prediction}: forecasting future trends using the temporal axis of the data; and 3) \textit{Segmentation}: dividing imagery into classes at the pixel level (semantic), object level (instance), or both (panoptic) \citep{kirillov2019panoptic}.

Cropland remote sensing exploits all three of these applications. Analysing past, present, and predicted crop distributions and yields from SITS data is now a key tool for governmental and non-governmental organisations to monitor food security, agricultural sustainability, and environmental change \citep{SITSYieldPredSecurity}. A prime example is the European Space Agency’s Sentinels for Common Agriculture Policy (\citep{sen4cap}, Accessed May 2026), which aims to build high‑volume SITS data pipelines from Sentinel‑1 \citep{sentinel1} and Sentinel‑2 \citep{sentinel2,sentinels} for crop type mapping, vegetation health assessment, and agricultural practice monitoring. These products support implementation of the European Union’s Common Agricultural Policy (\citeauthor{CAP}, Accessed May 2026), especially verification of per‑parcel reported crop type for allocating farming subsidies \citep{cropIDCAP}. When considering that  farmers’ average income is well below overall mean income, such tools must be as accurate as possible in both crop type and parcel boundaries to ensure fair subsidy distribution. While other cropland remote sensing methods exist, such as low‑altitude unmanned aerial vehicles \citep{UAVCrops} and traditional aerial imagery \citep{cropMappingSources}, satellite data offer unmatched spatial and temporal coverage at favourable cost, making SITS data the logical focus for large‑scale crop type mapping.

Supervised semantic segmentation of cropland has seen a particular influx of growth in recent years, in both models and datasets, largely due to the increasing availability of valuable ground truth data. Although performance boundaries are being continually pushed by novel architectures in this field, agricultural parcel boundary segmentation in particular remains challenging, with neighbouring parcels commonly seen either to be growing the same crop, or to be separated only by sub–spatial-resolution features such as fences \citep{nzCropBoundariesMethods}. The growth in datasets presents further challenge. Many factors in dataset design lack standard practice or consensus, and with little study on the effects of these choices, the capabilities of contemporary models in the field may be undersold by flawed dataset constructions.

This study examines and builds on these contemporary crop segmentation models and datasets. We firstly introduce our SITS crop segmentation model using a novel architecture of fully-factorised temporal-spectral-spatial attention, and also evaluate the continued relevance of convolutional components in transformer-heavy architectures through the addition of a hybridised convolutional refinement head. We then compare our model with several state-of-the-art models in the field, and analyse how informed crop-class selection and tile size choices in dataset design affect achievable segmentation performance. The key contributions of our work therefore follow as:

\begin{itemize}
     \item The introduction of the first crop segmentation model to apply factorised self-attention jointly over spatial, spectral, and temporal dimensions in SITS data. We propose a novel parallel encoder architecture for the temporal and spectral branches, both to achieve computationally viability of this approach, and to widen the reception of raw unattended signals from the original SITS inputs to a greater proportion of the network.

     \item Evidential findings to support the continued relevancy of hybridisations between attention and convolutional mechanisms in segmentation networks, improving performance significantly by leveraging the proven strengths of CNNS for spatial representations against the common trend of employing purely attention-based architectures.

     \item A comparative study of SOTA SITS crop segmentation models across an unprecedented coverage of tile-size variants from multiple datasets, demonstrating that tile size, even within the same dataset, strongly affects results and prevents fair comparison between models trained with different tile sizes. Our results also show that flawed class selections and aggregations of multiple crop species can degrade crop segmentation performance to a greater extent than model architectural choices, highlighting the importance of considerate practices for involved crop classes in construction of future datasets.

     \item We show our PAtteRNS model for crop segmentation to outperform existing SOTA models in key segmentation metrics, particularly in terms of parcel boundary quality, demonstrated via Boundary IoU as a measure of fine parcel delineation. We make our model publicly available for use and adaptation.
\end{itemize}

\section{Background}\label{sec:background}

\subsection{Basis of Deep Learning in Crop Segmentation}\label{sec:background:earlyModels}

Early ML efforts in cropland classification relied on hand-crafted spectral, temporal, and textural features, which were subsequently classified using models such as Support Vector Machines (SVMs), Decision Trees (DTs), and Random Forest (RF) \citep{palRandomForest, BelgiuRandomForest}, or through unsupervised methods such as K-means clustering \citep{AgricultureDL1028}. The shift in modern cropland classification to deep learning (DL) followed the popularisation of convolutional neural networks (CNNs) for vision tasks \citep{lecunCNN, alexNet, mortensen2016semantic, agriDLSITSreview}, whose ability to directly learn image feature representations suited the growing scale of available SITS data. While conventional CNNs were primarily designed for image-level classification tasks, the use of fully-convolutional networks (FCNs) then enabled dense per-pixel classification by replacing fully connected output layers with convolutional operations, making semantic segmentation more practical for agricultural imagery \citep{longFCNs, dyrmann2016AgriFCN, zhu2017DLRemoteSensing}. U-Net \citep{unet}, originally introduced for medical image segmentation, became one of the most popular early FCNs for segmentation tasks, with a proposed symmetric encoder-decoder architecture, along with skip-connections at each scale of representation to propagate finer spatially detailed information from encoder layers to the decoder. This architecture required comparatively limited labelled training samples while improving overall model speed and accuracy, particularly in the preservation of fine class borders, such as agricultural parcels, and as such saw rapid uptake within cropland segmentation \citep{resunetFields, UAVCrops, sen2UNettemporalcrop}. DeepLabV3+ \citep{chen2018deeplabv3plus} has seen more recent use in cropland segmentation \citep{deeplabv3landsat, sen2UNettemporalcrop} and represents a more recent and higher-performing evolution of this core concept. This is primarily achieved through atrous convolutions and spatial pyramid pooling allowing for greater receptive fields \citep{chen2017deeplabv3}, and a ResNet \citep{he2016resnet} backbone to enable efficient deep feature extraction through residual connections. These methods are limited to single temporal state images and do not jointly learn spatial and spectral information within meaningful temporal features. A common solution is to hybridise CNNs with a temporal network, typically a Recurrent Neural Network (RNN) such as Long Short-Term Memory (LSTM) \citep{hochreiterLSTM} or Gated Recurrent Units \citep{choGRU}, enabling learning of deep phenological crop features. These methods improved model performance, with implementations including CNN-LSTMs \citep{finneganSupeVSUnsupe}, ConvLSTM \citep{shiCNNLSTM}, and Bi-ConvGRU \citep{mtlcc}. Some stand-alone RNN and LSTM models outperformed CNNs due to their multi-temporal reasoning ability \citep{russwurm2017rnnlstm}. Alternatively, instead of using RNNs to model temporal information, convolutional architectures have been directly extended to higher dimensions. 3D CNNs \citep{3DCNN} and U-Net3D \citep{unet3d} treat satellite image sequences as volumetric input, enabling joint learning of multi-dimensional features via 3D convolutions.

\subsection{Introduction of Vision Transformers}\label{sec:background:transformers}

The Vision Transformer (ViT) \citep{vit} significantly advanced numerous computer vision tasks \citep{rsSegmentationSurvey2023, arnab2021vivit}. Based upon self-attention mechanisms \citep{attentionIAYN}, the application of the Transformer architecture to images was achieved through breaking them into tokenised non-overlapping patches, allowing for computation in the same fashion as tokens would receive in a natural language processing model. The basic structure of a ViT encoder is formed through repetitive layers of multi-head attention and feed-forward components, joined at each interface with a layer norm and a residual connection (Figure 2). These ViT models achieved excellent results at reasonable computational cost, but require far more labelled training data than prior methods. Initially proposed for classification, the ViT was soon extended to semantic segmentation with the Pyramid Vision Transformer (PVT) \citep{wang2021pyramid}, which introduced a hierarchical feature pyramid for multi-scale representations while reducing the cost of standard self-attention, demonstrating the suitability of pure-transformer backbones for dense prediction tasks. Notable subsequent works included Swin Transformer \citep{liu2021swin} and SegFormer\citep{xie2021segformer}, both attempting to further improve the computational viability of vision transformers. ViT offers global attention over the whole image but has quadratic complexity in the input sequence length. Swin Transformer reduces this to linear complexity by computing self-attention only within local patch regions, then mixing these regions in later encoder layers to approximate global attention. In contrast, SegFormer showed stronger performance by combining local and global attention by pairing a multi-scale transformer hierarchical encoder and a lightweight multilayer perceptron (MLP) decoder. Another key innovation in the ViT architecture was the classification $cls$ token introduced in BERT \citep{devlin2019bert}. Prepended to the embedded token sequence in transformer models, this single learned token was shown to suffice for classification, replacing the need to pool all tokens and thereby reducing model size while often improving accuracy. 

\begin{figure}[htb]
\centering
\includegraphics[width=0.2\textwidth, keepaspectratio]{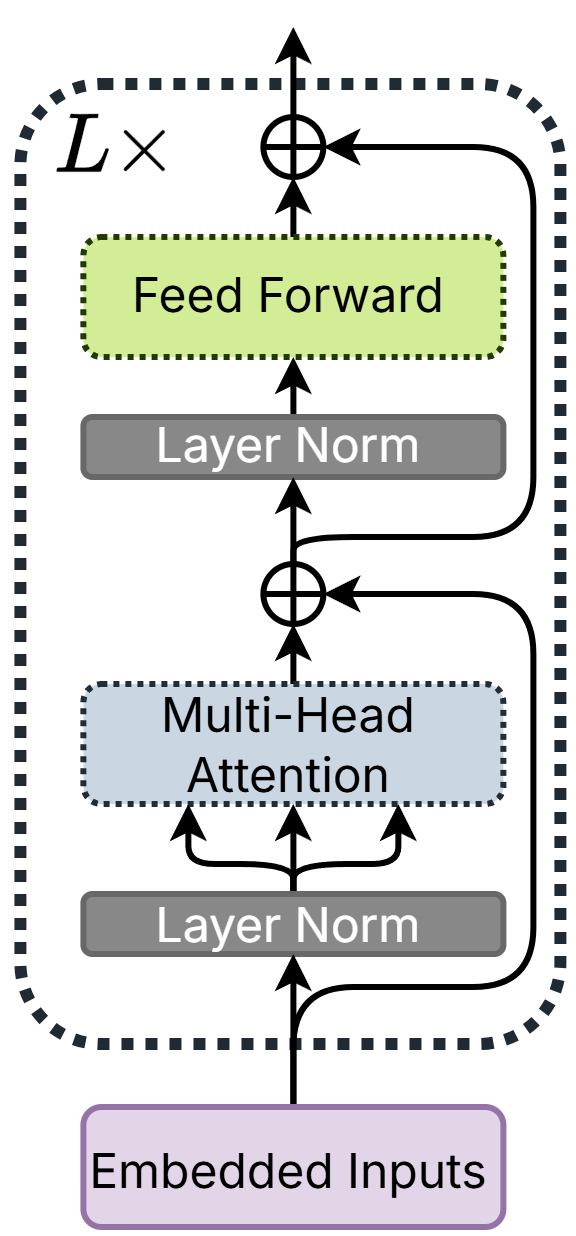}
\caption{\textbf{The structure of a vision transformer encoder block.} This structure is repeated to the desired depth of the transformer $L$}
\label{fig:vitencoder}
\end{figure}

While ViT operates in a two-dimensional space, its applicability can be extended to higher dimensions using ViViT \citep{arnab2021vivit}, which encodes video by extracting non-overlapping spatio-temporal tubelets and projecting them into embedded token sequences that jointly capture spatial and temporal aspects of the input. A key challenge in linking and lengthening token sequences is the quadratic complexity of self-attention. To address this, ViViT factorises self-attention in its transformer backbone, computing spatial attention then temporal attention sequentially over the embedded tokens, enabled by simple reshaping of the tokens before each attention step. The results showed that factorised self-attention is a strong approach for modelling multi-dimensional interactions. However, cross-attention can also do this \citep{attentionIAYN, gheiniCrossAttn, wen2023crossattention}, by using queries from one token sequence and keys/values from another. This lets one sequence directly influence another and can be applied to spatial and temporal representations \citep{crossAttentionMapsSegmentation}.

\subsection{Contemporary Crop Segmentation Models}\label{sec:background:newModels}

Models tackling SITS crop segmentation problems have been shown to benefit greatly from the specific inclusion of multiple temporal samples \citep{zhu2017DLRemoteSensing, temporalCNN, pastis}, aided by improved temporal resolution and range in SITS data and by reduced data storage and processing costs \citep{mtlcc}. SITS sources with revisit periods under 7 days are now commonly and openly available \citep{sentinel2, PlanetLabs}, providing over 50 temporal observations each year. Aside from external constraints on method choice, models that ignore temporal data for crop segmentation are usually passed-by in contemporary works. Similarly, although unsupervised \citep{unsupervised} and semi-supervised \citep{semisupervised, finneganSupeVSUnsupe} methods for SITS cropland segmentation have seen past popularity by reducing the need for extensive ground truth labels, improvements in ground truth label availability (Section \ref{sec:background:datasets}) now see supervised models as the dominant choice for crop mapping in many areas. Leveraging temporal features, two seminal recent supervised crop segmentation models both use self-attention to process the temporal data axis.

The first of these, UTAE \citep{pastis}, combines a U-Net style CNN encoder-decoder architecture with attention mechanisms. Spatial relationships are handled by CNNs, and temporal dependencies are extracted at the lowest convolutional resolution by a lightweight transformer encoder. Temporal attention masks are interpolated upwards across all other resolutions levels, avoiding the cost of high resolution temporal encoding while still providing full temporal attention. Introducing the temporal attention mechanism achieved state-of-the-art (SOTA) performance over all prior non-attention models.

The second of these models, TSViT \citep{tsvit}, removed all convolutional components, using a purely transformer-based architecture with a lightweight MLP head and per-dimension attention factorisation similar to ViViT. While ViViT found no difference in whether temporal or spatial attention came first in video encoding, TSViT shows a clear benefit to placing the temporal encoder before the spatial encoder, and argues this order is intuitive for cropland segmentation because per-crop phenology is highly discriminative. TSViT reports state-of-the-art performance across several benchmark datasets.

Although both models use temporal attention to varying degrees, we find that attention mechanisms along the spectral axis of SITS products remain relatively under-studied, even though indices derived from different spectral bands are a long-established tool in multispectral remote sensing \citep{thenkabail2012hyperspectral, pena2011objectindicies, boyd2002RS}. Only a few recent works address spectral attention. Two studies apply transformer-based segmentation networks to spectral-spatial imagery for crop segmentation or classification, but neither factorises attention between spatial and spectral. One of them mainly compares different spectral band sets \citep{spectralSpatialTransformerDatePalms}, the other uses various CNNs for feature extraction before the transformer encoder \citep{spectralSpatialTransformerCorn}, and both target a single crop species. Another study separates attention across different spectral bands before a cross-attention fusion \citep{vitCrossSpectralAttention}, but lacks a temporal axis, does not use SITS, and employs only a few spectral bands. A recent work, SSF-TransUnet, factorises attention into spectral and spatial components for crop classification and uses separate ResNet-based feature-extraction branches for these two axes, but still lacks a temporal axis. That study uses $1m$ resolution imagery from one satellite for the spatial branch and $10m$ resolution imagery from another satellite for the spectral branch, which limits the method’s broader adaptability. To our knowledge, no prior work has used a fully factorised attention approach with spectral attention for SITS crop segmentation, despite the strong performance of factorised attention in multi-dimensional transformers in recent years.

Recent works show that modern DL architectures, including RNNs \citep{mtlcc, cropMappingHeirarchyDL} and attention mechanisms \citep{selfattnSITS, pastis}, can jointly learn the masking of cloud alongside crop type classifications in SITS segmentation tasks. This significantly lowers the entry barrier for SITS segmentation and allows models to achieve excellent results without explicit cloud-handling procedures. A notable omission in most reviewed crop class segmentation studies is an assessment metric for parcel boundary quality, despite its importance for verifying self-reported parcel boundaries in crop subsidy and land management systems. While segmentation methods can reach high performances using traditional segmentation metrics such as Overall Accuracy (OA) or Mean Intersection over Union (mIoU), these can fail to represent how consistent a model performs in boundary pixel placement \citep{cropBoundariesReview2026}. Several works focus specifically on agricultural boundary delineation, rather than generalised crop class segmentation, yet none converge on any single ideal metric to quantify boundary accuracy \citep{nzCropBoundariesMethods, resunetFields}. This indicates that SITS crop mapping has not yet adopted a widely suitable metric for evaluating boundary quality in segmentation results.

\subsection{Contemporary Crop Segmentation Dataset Landscape}\label{sec:background:datasets}

For many years the NASA Landsat Series of satellites was a primary  source for crop type mapping \citep{boyd2002RS, deeplabv3landsat}. More recently, Sentinel-2 and numerous commercial satellites have been used more often \citep{flairDatasetCrops, agriDLSITSreview}, greatly increasing the volume of available SITS raw data and facilitating accessibility of SITS data analysis as well as the production of cropland semantic segmentation datasets \citep{datasetsreview, sykas_sen4agrinet_2021, denethor} based on both public and commercial satellite imagery. For many of these, the need for ground truth labels is met by public access to government databases of crops in each parcel, typically those used for subsidy control \citep{agriDLSITSreview}. This single source type has generated many high-quality supervised learning datasets with large numbers of labelled crop parcels, suitable even for data-hungry architectures \citep{eurocrops}. However, this diversity has also produced inconsistent practices, including translation errors in crop class names, aggregate classes containing different sub-crops across datasets, and uneven recognition of crop seasonal variants \citep{EUCropHCAT}.

Generally, naming errors are traceable, independent of method results, and therefore not covered here. Some crop classes, even when taxonomically well defined, have less homogenous ground coverage than others as seen with `Meadows', which contains many individual coverage species, or orchard and vineyard-based crops, where the primary species is interspaced with grasses and other ground coverage. As these are still valid taxonomical classifications, we term these classes as `Non-Homogenous'.

A more common problem is how crop class taxonomies are chosen and grouped. Some datasets merge seasonal variants of a crop (e.g., `Rapeseed') into a single class \citep{mtlcc}, while others distinguish `Winter Rapeseed' \citep{pastis} but omit the `Spring' and `Summer' variants \citep{EUCropHCAT}, making the true overlap of these crops across datasets unclear. We term this grouping practice as `Seasonal Aggregate' classes. Elsewhere, broad aggregate labels group taxonomic and phenological differences across multiple crops, ignoring heterogeneity in spectral and temporal features and its potential impact on model performance. For instance, the popular PASTIS dataset groups fruits, vegetables, and flowers as one class \citep{pastis}. We term these groupings as `Taxonomical Aggregate' classes. These inconsistencies mean a class boundary in one dataset may not correspond to any meaningful boundary in another, complicating accuracy comparisons across studies and potentially inflating or deflating reported performance depending on how forgiving or strict the underlying taxonomy is. These groupings may be especially harmful in long-term studies,  preventing models from using region-specific planting windows.

Another often seen flaw is the inclusion of only a few dominant classes within a study or dataset \citep{unet3d, sen2UNettemporalcrop, unsupervised, cropMappingSources, spectralSpatialFactorisedCropTransformer}, omitting minority classes, which may still account for a large share of cropland in the study area. This is often done to balance class populations, however this artificial boost in model performance may not reflect real world deployment capabilities \citep{cropMappingHeirarchyDL}.

The final aspect of the dataset standardisation reviewed is the choice of tile size. Datasets use a wide range of sizes, in both pixel dimensions and spatial extent. We find no studies that explicitly analyse how tile size in constructed crop segmentation datasets affects model performance, or that compare a single model’s performance across multiple tile sizes.

\section{Methods}\label{sec:methods}

\subsection{Datasets}\label{sec:methods:data}

\begin{figure}[htb]
\centering
\includegraphics[width=0.48\textwidth]{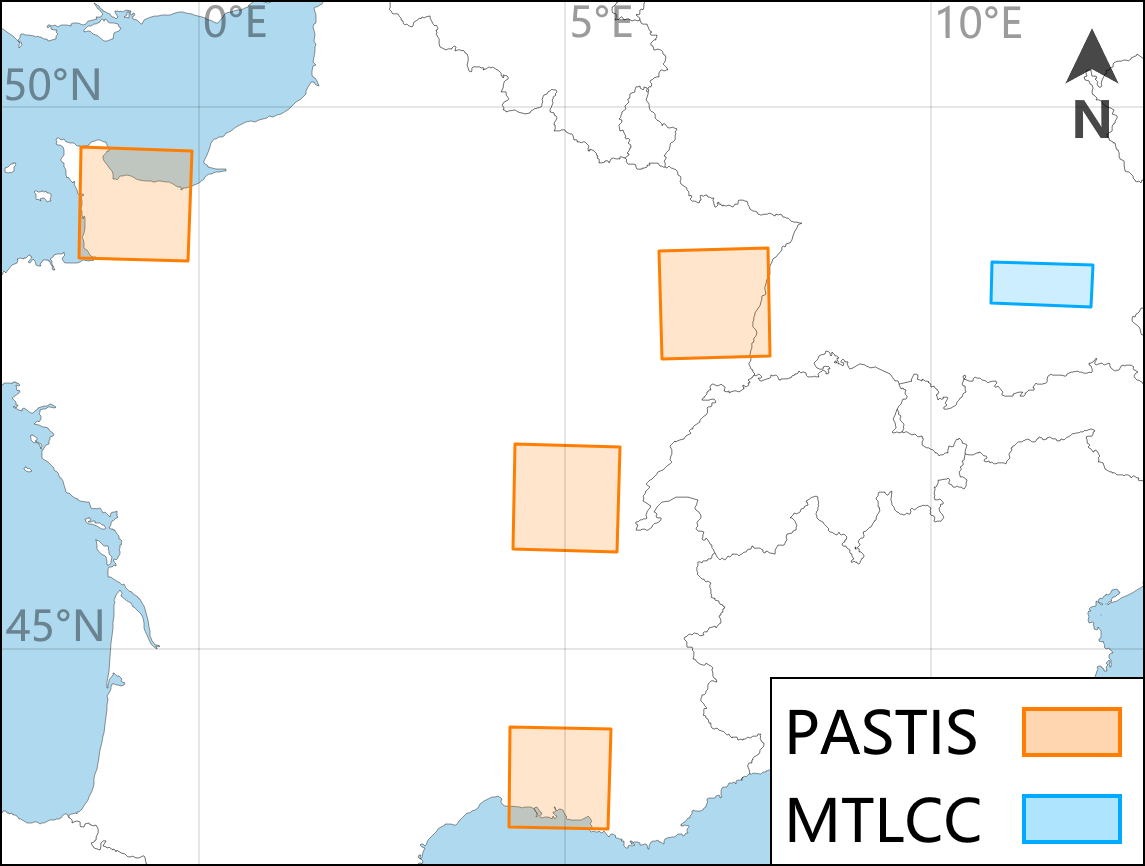}
\caption{\textbf{The geographical extent of the two datasets selected for this study across France and Germany.}}
\label{fig:datasetMap}
\end{figure}

We selected datasets with predefined tile-size variants and similar class counts, localities, and temporal ranges, favouring those with many crop classes, temporal samples, and data points to better assess real-world performance. This led to using two Sentinel-2-derived datasets, each with two previously used or original tile-size variants. The first dataset, PASTIS \citep{pastis}, covers French agricultural parcels across 4 separate Sentinel-2 full-size tiles, with data provided by the French Land Parcel Identification System. The second dataset, MTLCC \citep{mtlcc}, covers a much smaller geographical spread within Germany, but has a denser coverage of parcels within this area, provided by the Bavarian Ministry of Food, Agriculture and Forestry. Figure \ref{fig:datasetMap} shows the spatial extent of the two datasets.

Table \ref{tab:datasets} gives the characteristics of both Datasets. For both datasets, the Sentinel-2 $20m$ and $60m$ spatial resolution bands were bi-linearly interpolated to the $10m$ resolution for algorithmic simplicity and compatibility. PASTIS has a single temporal variant, running from September 2018 to November 2019, and MTLCC has two temporal variants, 2016 and 2017. Because the MTLCC ground truth labels differ between years and are not cross-compatible, we used the 2016 variant for its larger sample size. Both datasets contain cloudy data in some temporal bands of each tile, adding challenge and realism to model testing. The total amount of crop classes (Table \ref{tab:classes}), is similar between datasets, although with only partial overlap, allowing for a greater evaluation of the performance of the model against a variety of crops.

\begin{table*}[ht!]
\begin{adjustbox}{max width = \textwidth}
\begin{tabular*}{\textwidth}{@{\extracolsep\fill}lcccc}
\toprule
& \multicolumn{2}{c}{PASTIS} 
& \multicolumn{2}{c}{MTLCC 2016} \\
\cmidrule(lr){2-3}\cmidrule(lr){4-5}
Region & \multicolumn{2}{c}{France} & \multicolumn{2}{c}{Germany} \\
Sentinel-2 Included Band Sizes & \multicolumn{2}{c}{10m, 20m ($S=10$)} & \multicolumn{2}{c}{10m, 20m, 60m ($S=13$)} \\
Capture Date Range & \multicolumn{2}{c}{406 Days} & \multicolumn{2}{c}{341 Days} \\
Max Temporal Samples $T_{max}$ & \multicolumn{2}{c}{61} & \multicolumn{2}{c}{45} \\
Total Classes & \multicolumn{2}{c}{20} & \multicolumn{2}{c}{18} \\
Crop Classes & \multicolumn{2}{c}{18} & \multicolumn{2}{c}{17} \\
Folds & \multicolumn{2}{c}{5} & \multicolumn{2}{c}{10} \\
Nominal Train/Val/Test Split \% & \multicolumn{2}{c}{60/20/20} & \multicolumn{2}{c}{60/20/20} \\
\midrule
Tile Size & $128px$ & $24px$ & $48px$ & $24px$ \\
Total Used Tiles & 2,435 & 60,825 & 8,125 & 43,963 \\
Total Spatial Pixels (M) & 39.9 & 35.0  & 18.7 & 25.3 \\
Total Data Points (B) & 24.3 & 21.4  & 11.0 & 14.8 \\
\bottomrule
\end{tabular*}
\end{adjustbox}
\caption{\textbf{Properties of the datasets and tile-size variants used in this study.}}
\label{tab:datasets}
\end{table*}

These datasets provide 4-dimensional SITS data, characterised per image tile $I$ as $T_{temporal} \times S_{spectral} \times H_{height} \times W_{width}$. For tiles with $T < T_{max}$, we padded the temporal axis to length $T_{max}$ to support models requiring a fixed temporal extent. Both datasets show some of the crop class grouping issues described in section \ref{sec:background:datasets}. Out of the 18 crop classes in PASTIS, we flag 4 as taxonomical aggregates, making up 11.2\% of the primary crop parcels in the dataset, but only 4.1\% of crop class pixels. However, by the nature of the taxonomical aggregate grouping, these also contribute to non-homogenous ground coverage alongside the `Meadow' and `Grapevine' classes, bringing the percentage of non-homogenous crop parcels in the dataset up to 60.0\%, and pixels to 25.2\%. PASTIS does not exhibit any seasonal aggregate groupings within it's classes. In MTLCC, only one of its 17 crop classes is a taxonomic aggregate: `Winter Wheat`, combining soft and hard winter wheats \citep{EUCropHCAT}, yet it still accounts for 24\% of primary crop parcels and 14.1\% of pixels. In combination with `Meadow', MTLCC reaches a 37.6\% by-parcel and 17.9\% by-pixel coverage of non-homogenous crops. MTLCC also features a potential seasonal aggregate class of `Rapeseed' over 3.8\% of parcels and 2.5\% of pixels, as opposed to the `Winter Rapeseed' crop class used by PASTIS. It is unclear what mix of spring, summer, and winter rapeseed varieties this aggregate class covers \citep{EUCropHCAT}.

Both datasets include a `Background` class alongside their main crop classes, but the PASTIS dataset also introduces a `Void` class, which is an additional concern. Any crop parcel in the original ground truth whose area lies more outside the tile than inside, or whose crop class is not in the PASTIS dataset, is labelled as `Void'. This differs from `Background` and introduces several issues, mainly that void crop parcels within a tile can confuse learning and inflate a model’s reported false positive rate due to potentially containing otherwise valid crop classes. Consequentially, ground truth parcels of the `Void' class are masked during learning and metric evaluations for PASTIS, nullifying the effects of this class on the models. Differing in approach, in MTLCC, crops not included in the label set but present in a tile remain labelled as `Background`. Models trained on PASTIS thus have a theoretical performance advantage, as parcels of non-included classes are removed, reducing label confusion. Merging the `Void` and `Background' classes in PASTIS could be considered, but it would still leave a proportion of otherwise-valid crop class parcel segments at tile boundaries classified as `Background' and so retain the issue of inflated crop class false positives. Our decision not to mask the `Background' classes is motivated in Section \ref{sec:methods:training}.

Using a unified hierarchical taxonomy of crop nomenclature in EU states \citep{eurocrops, EUCropHCAT} and the French LIPS nomenclature \citep{rpg_france}, the class names of both datasets have been standardised where they intersect. The complete combined class set along with labelling of per-class aggregation and homogeneity issues is given by Table \ref{tab:classes} in \ref{app:datasetClasses}.

\subsection{Baseline Transformer Crop Segmentation Model}\label{sec:methods:transformers}

\begin{figure*}[ht]
\centering
\includegraphics[width=0.8\textwidth]{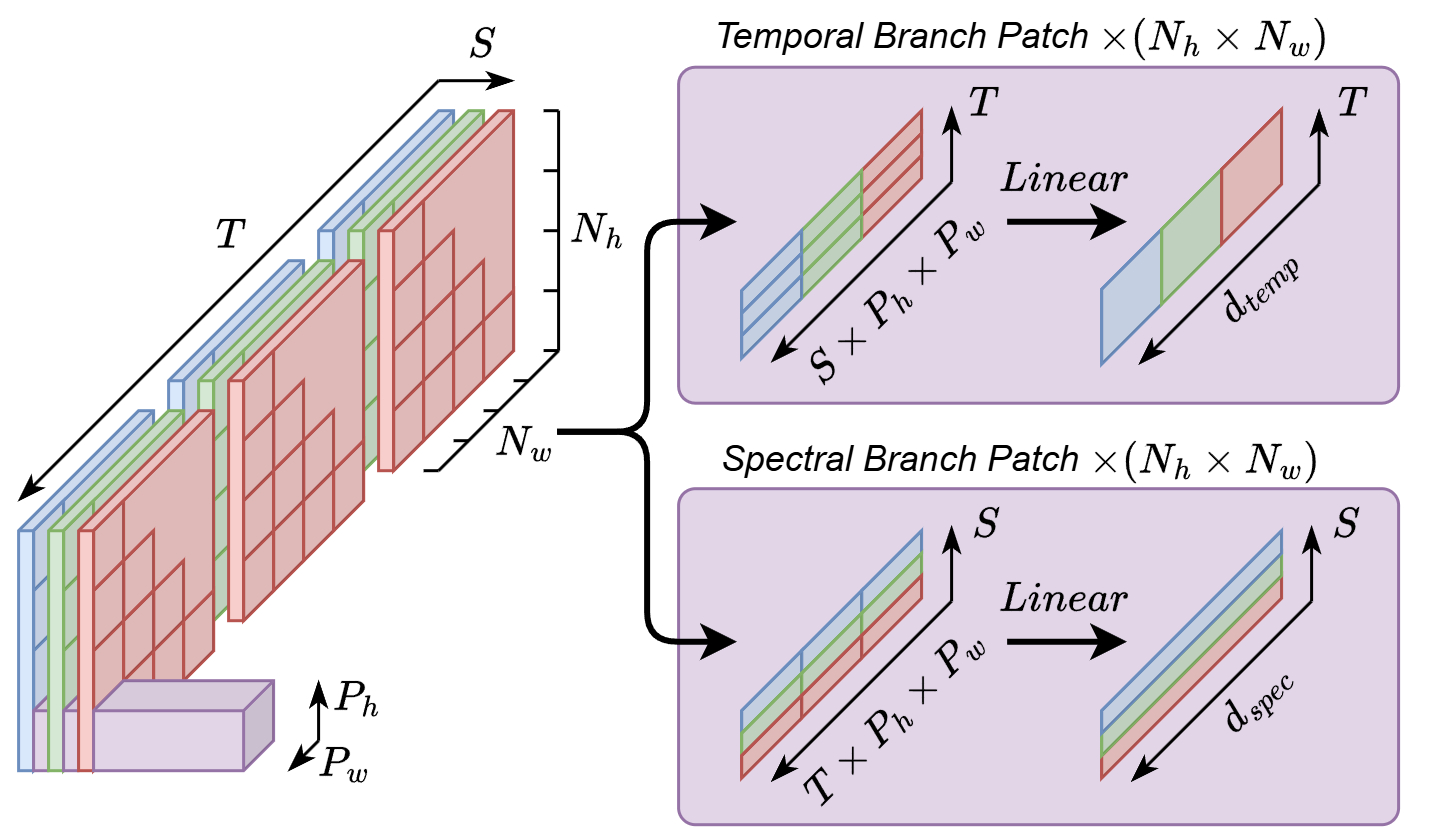}
\caption{\textbf{The parallel temporal and spectral patch extraction structure from raw SITS data.} The input SITS tile tensor $I\in\mathbb{R}^{T\times S\times H\times W}$ is taken twice, once for the temporal transformer branch (top), and once for the spectral transformer branch (bottom). Each branch begins by partitioning the raw data into the same pattern of non-overlapping spatial patches with size $P_h\times P_w$, giving $N_h=H/P_h$ and $N_w=W/P_w$ patches. Each branch then constructs a token sequence per patch, with the temporal branch flattening each patch along the spectral and spatial dimensions $S \times P_h \times P_w$ while preserving the temporal dimension, and the spectral branch flattening each patch along the temporal and spatial dimensions $T \times P_h \times P_w$ while preserving the spectral dimension. This yields $N = N_h \times N_w$ token sequences per each of the spectral and temporal branches, which are separately projected through a linear layer into the embedding dimension of size $d$.}
\label{fig:patching}
\end{figure*}

Our model is implemented in PyTorch \citep{pytorch}, using an efficient scaled dot-product multi-head attention implementation (\citeauthor{pytorchSDPA}).

For a baseline model architecture, due to the success of TSViT \citep{tsvit}, we find no reason to deviate from its temporal-then-spatial factorised attention layout with a ViT-based backbone \citep{vit}. The backbone consists of multiple identical transformer encoder blocks (Figure \ref{fig:vitencoder}), each adapted to specific dimensionalities by shaping their input data. A single tile of SITS input is a tensor $I\in\mathbb{R}^{T\times S\times H\times W}$ where $T$ represents the temporal sequence length, $S$ the number of spectral bands, and $H$ and $W$ are the original tile's height and width in pixels. This is firstly split into $N$ non-overlapping spatial patches of size $P$ thus producing $(H \times W)/P^2$ patches . Each patch is flattened by linearly projecting its spectral-spatial dimensions $S \times P^2$ into an embedding of size $d$ while preserving time, producing a token sequence $\mathbf{x} \in \mathbb{R}^{N \times T \times d}$ representing the tile.

Our embedded patches are augmented with positional embeddings $pos$ and $cls$ tokens. Temporal positional encodings use an embedded normalised value representing each capture date’s relative position between the sequence start and end dates(details in section \ref{sec:methods:dates}). A $cls$ token is a learnable $d$-dimensional embedding that serves as a fixed-size summary of a tile; at the output, its state directly represents the predicted pixel classes. Following \citep{tsvit}, we use $k$ and $cls$ tokens as the baseline representation size, where $k$ is the number of dataset classes, giving $cls \in \mathbb{R}^{k \times d}$. After these additions, a token sequence consists of concat$(cls, \mathbf{x}+pos) \in \mathbb{R}^{N \times (k+T) \times d}$. This token sequence is passed to the temporal transformer, which then operates over each of the $N$ patch positions, attending across the $T$ temporal tokens at that position to model the temporal dynamics of that patch over the observed sequence. After the temporal transformer, we keep the $k$ and $cls$ token representations at each patch position and discard the other temporal outputs, resulting in a tensor of shape $N \times k \times d$. The $N$ patch-level representations, each with $k$ per-patch $cls$ tokens, are flattened and projected through a learned linear layer into the spatial embedding space, with a matching embedded dimension size to the temporal embedding. A learned spatial positional encoding $pos \in \mathbb{R}^{N \times d}$ is added to the projected sequence before it enters the spatial transformer, which attends over all $N$ patches so each patch can incorporate contextual information from the tile. The output of the spatial transformer undergoes an `unpatch' operation, expanding the raw outputs of shape $N \times d$ via a linear layer to $N \times (P^2 \cdot k)$, which is then rearranged back to the tile's original spatial arrangement at full pixel resolution. The produced output tensor of shape $k \times H \times W$ is suitable for $argmax$ classification on a per-pixel basis.

This baseline factorised transformer crop segmentation model now forms the starting block for our proposed model. Because our transformer model requires token sequences of equal length, we pad samples with fewer image capture dates along the temporal axis with 0. We opted to pad rather than interpolate to not introduce unknown biases into experiments. Pad-masking is used in the attention mechanisms of our architecture to track these padded temporal bands. Before changing the architecture, the main way to improve performance in this vision-transformer model is to adjust the patch size $P$. Reducing $P$ has been shown to improve ViT-based models on dense prediction tasks \citep{xie2021segformer, tsvit}, but it directly increases the computational cost for a given model. Therefore, we do not fix the operating patch size in the architecture; instead, we keep it task-dependent and evaluate multiple patch sizes to assess the costs and benefits of this refinement.

\subsection{Proposed Parallel Temporal-Spectral Attention Structure}\label{sec:methods:parallelspectral}

When proposing a spectral data axis equal in importance and implementation to the temporal component of the baseline factorised model, an initial intuition is to arrange the temporal, spectral, and spatial components sequentially. In contrast, we propose a factorised architecture with parallel spectral and temporal components, each processing patched representations of the original SITS data, followed by the spatial transformer operating on embedded representations already attended by each parallel primary transformer block. This architecture is illustrated via Figure \ref{fig:patching}.

Instead of sequentially processing temporal and spectral dimensions, where the second transformer attends to a representation already processed by the first, our dual-tokenisation strategy lets the model learn temporal and spectral dependencies independently from the same spatial patches, so both transformers operate in parallel on the original embedded SITS data.
In contrast, a sequential architecture that retains high-dimensional token representations between transformers produces very large intermediate activations. Since self-attention scales quadratically with sequence length, this greatly increases computation and memory relative to our parallel design; we have explored the sequential architecture and observed an approximate 5-fold increase in multiply-accumulate operations (MACs) for the sequential model without token reduction between transformers. Restricting the intermediate representation to class tokens cuts this cost but may remove information, especially for the third transformer in the sequential model which would receive twice-compressed token representations, limiting it's ability to model complementary relationships.

The spectral branch of our architecture is handled largely the same as the baseline’s temporal section. The spectral branch receives one of the two copies of each $N=N_h \times N_w$ non-overlapping spatial patches, collapsing the temporal and spatial dimensions of each patch to produce a vector of size $S \times (T \times P^2)$ (Figure \ref{fig:patching}). A linear projection into the shared embedding dimension size of $d$ then produces a spectral branch token sequence of shape $\mathbf{x}_\mathrm{spec} \in \mathbb{R}^{N \times S \times d}$. The temporal branch is unchanged from the baseline, collapsing instead the spectral and spatial dimensions in each of the $N$ patches to give a temporal branch token sequence of shape $\mathbf{x}_\mathrm{temp} \in \mathbb{R}^{N \times T \times d}$.

The difference between the spectral and temporal branches outside of the patching operation lies in the added spectral positional encodings $pos_{\mathrm{spec}}$, which, like spatial positional encodings, are learned tokens $pos_{\mathrm{spec}} \in \mathbb{R}^{S \times d}$, rather than the SITS capture date embeddings used in the temporal branch. We add $pos_{\mathrm{spec}}$ tokens to the embedded sequence before concatenating the $k$ and $cls$ tokens, as in the temporal branch, producing concat$(cls, \mathbf{x}+pos_{\mathrm{spec}}) \in \mathbb{R}^{N \times (k+S) \times d}$. This sequence is input to the spatial transformer, which attends over the $S$ spectral bands for each of the $N$ embedded patches. As in the temporal transformer, we then discard the original embedded tokens, retaining only the $cls_\mathrm{spec}$ tokens of shape $N \times k \times d$.

With both parallel transformer branches producing the equally-dimensioned $N \times k \times d$ sets of $cls$ tokens, $cls_\mathrm{temp}$ and $cls_\mathrm{spec}$, the spatial transformer must receive a fusion of these sets. A simple concatenation is possible for this example, producing a token sequence of dimensions $N \times 2k \times d$, however more complex fuse block implementations are possible, as covered in section \ref{sec:methods:crossattn}. For the spatial transformer, the only required alteration to the baseline model is in the linear layer used to project the pre-attended token sequence into the shared embedding dimension $d$, which now receives the expanded $N \times 2k \times d$ input. The embedded representation going into the spatial transformer is represented as $\mathbf{x}+pos_{\mathrm{spat}} \in \mathbb{R}^{N \times d}$. After the spatial transformer, the original 'unpatch' operation is conducted, with a linear layer projecting to $N_w \times N_h\times \cdot k$, before rearrangement to the original tile spatial dimensions with per-class probability representation of $k \times H \times W$.

\subsection{Temporal-Spectral Cross-Attention}\label{sec:methods:crossattn}

\begin{figure}[htb]
\centering
\includegraphics[width=0.4\textwidth]{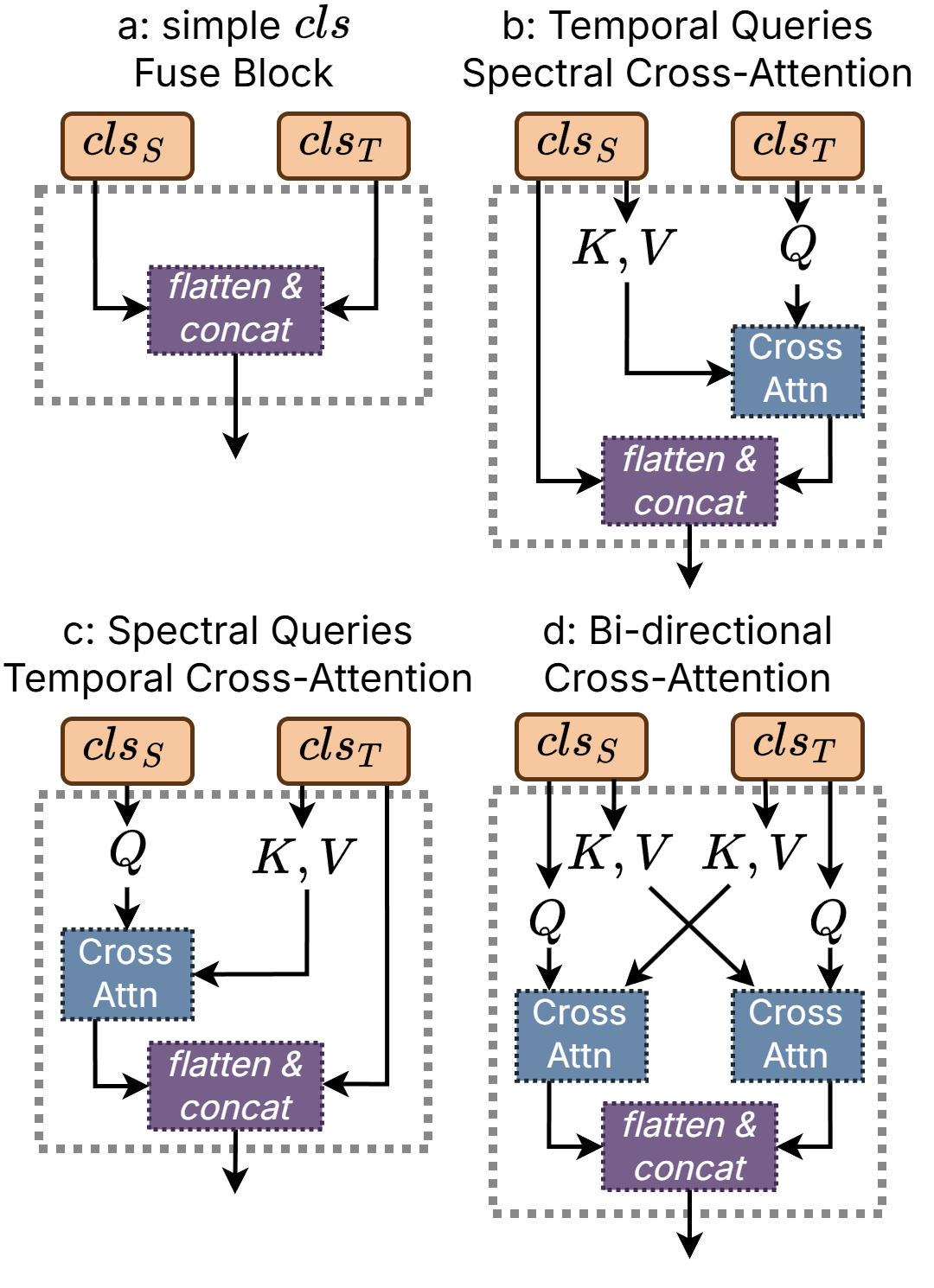}
\caption{\textbf{The compared Fuse Block structures for the $cls$ token fusion following our parallel spectral and temporal transformer branches.} The outputs of these fuse blocks are passed onwards to the spatial embedding process, and then the spatial transformer.}
\label{fig:fuseblocks}
\end{figure}

Because our model uses two parallel representation paths, the two sets of $k$ $cls$ tokens output by the temporal and spectral transformers must be merged. We implement and evaluate several cross-attention methods for this fusion block. Unlike self-attention, cross-attention takes its query and key/value pairs from different sources rather than the same sequence. Given one sequence as an originating query $Q$, and another as a contextual key/value pair $K,V$, the output is computed through standard scaled dot-product attention. This allows one representation to selectively pull information from the token sequence in a learned fashion, therefore allowing our spectral or temporal branches the opportunity to exchange information prior to fusion that could be used to represent complex spectral-temporal relationships. It is also possible in a paired input such as our architecture for both token sequences to attend each other in a balanced fashion. The four fuse blocks are shown in Figure \ref{fig:fuseblocks} and described below.

\begin{enumerate}[a:] 
    \item This is a simple concatenation-based fusion block for the $cls_\mathrm{S}$ and $cls_\mathrm{T}$ tokens.
    \item In this temporal queries spectral cross-attention block, $cls_\mathrm{T}$ serves as the query $Q$, and $cls_\mathrm{S}$ as the key–value pair $K, V$. The temporal representation attends to the spectral representation, selectively incorporating spectral context into the temporal tokens.
    \item This spectral queries temporal cross-attention block is the opposite of fuse block b, with $cls_\mathrm{S}$ used as $Q$ and $cls_\mathrm{T}$ as $K, V$. The spectral tokens now draw on the temporal tokens for additional representational context.
    \item Unlike blocks a, b, and c, this bi-directional cross-attention block lets both token sets act simultaneously as query and context, enabling mutual information exchange instead of one-way information flow.
\end{enumerate}

\subsection{Convolutional Refinement Head}\label{sec:methods:convhead}

\begin{figure*}[ht]
\centering
\includegraphics[width=0.95\textwidth]{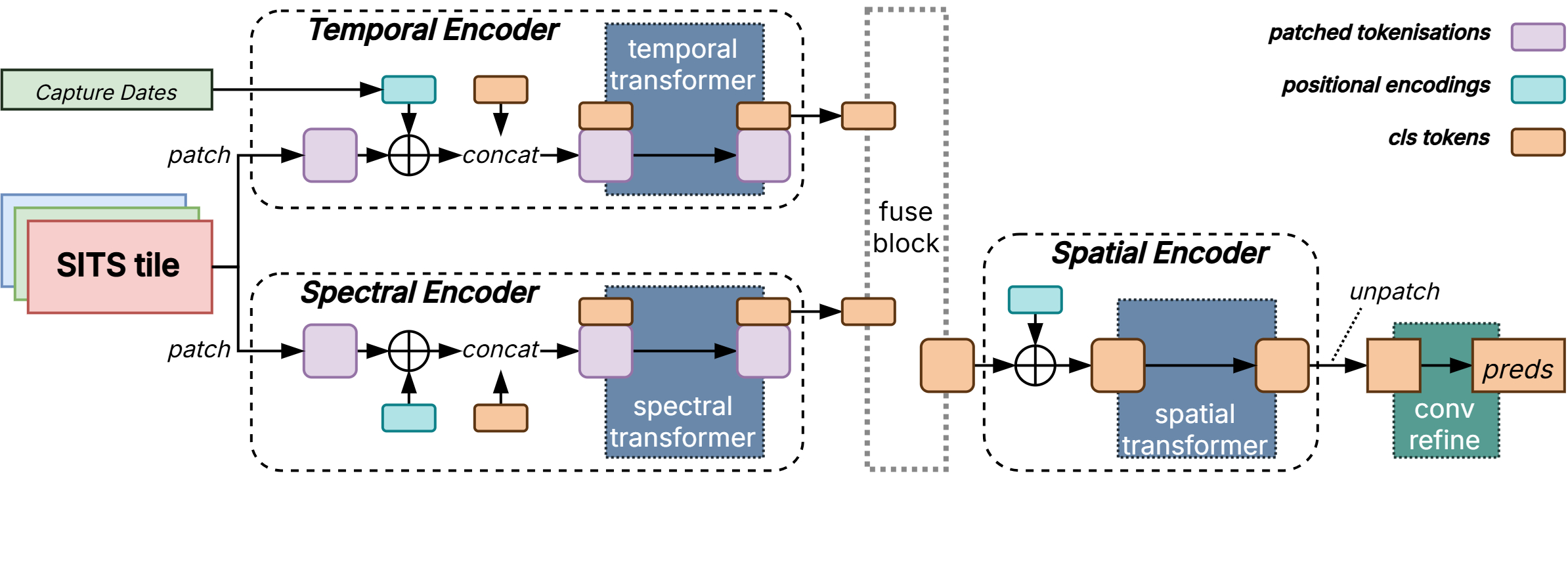}
\caption{\textbf{The overall architecture of our SITS Crop Segmentation Model.}
We name our model as PAtteRNS: \textbf{P}arallel \textbf{Atte}ntion and \textbf{R}efinement \textbf{N}etwork for \textbf{S}egmentation. While this network is intended for cropland segmentation, the same concepts may apply across other fields in remote sensing.}
\label{fig:wholearchitecture}
\end{figure*}

Although outputs from our spatial transformer block can be directly restructured into the original tile pixel layout, making the architecture purely attention-based, we are aware of the established strengths of CNNs for spatially complex tasks. We therefore add an optional Convolutional Refinement Head (CRH), which operates on the unpatched transformer outputs of shape $k \times H \times W$. Positioning the CRH directly before the model output (see Figure \ref{fig:wholearchitecture}) maximizes its portability to future architectures, including purely attention-based models like ours and more traditional models that already use convolutions. A small exploration of possible combinations of layers and convolution sizes was undertaken, while aiming to minimise any computational cost. Based on this, our proposed CRH uses three convolutions: a $3\times3$ layer that expands the $k$ input channels to $e = \max(32, 4k)$, a second $3\times3$ layer that processes this representation, and a final $1\times1$ layer that projects back to $k$ channels. The first two convolutions preserve the $H \times W$ spatial size via padding and are followed by batch normalisation and ReLU; the final $1\times1$ layer is left linear without normalisation or activation so that it can learn unrestricted corrections to the original $(k\times H\times W)$ output space. This residual is added element-wise to the spatial transformer’s unpatched output, so the CRH refines rather than replaces the class logits. Visualised in Figure \ref{fig:wholearchitecture}, our proposed architecture is now fully detailed.

\subsection{Temporal Encodings}\label{sec:methods:dates}

We observed two dominant date indexing methods in the literature. The first, Day of Year (DOY), poses the positional encodings by the given date within its calendar year, such that the minimum value is January 1\textsuperscript{st}, and the maximum is December 31\textsuperscript{st} with a pre-normalisation range of 0 through 365. The second, which we term Days since Reference Date (DSRD), uses either the exact first chronological capture date within the dataset, or a date just before the first capture as the minimum value for pre-normalisation capture dates. As an example, for PASTIS, a reference date of September 16\textsuperscript{th} 2018 is used, 1 day before the first capture, and a final capture date of October 27\textsuperscript{th} 2019 gives a range of 0 through 407.

As the selected datasets have limited multi-year coverage, extensive comparison between them would be out of scope; both are instead treated equitably within the single-year MTLCC data. This choice might have an effect on model performance within PASTIS, as its temporal span exceeds one year and its start date does not align with the start of a calendar year. We experimentally assess the impact of this potential performance factor in the ablation study reported in Section \ref{sec:results:architecture}. For all other experiments, we use DSRD encoding for our model to better standardise between PASTIS, with it's offset temporal span from a calendar year, and MTLCC, which covers the 2016 calendar year. For comparative models, we use the original implementation's chosen encoding method.

\subsection{Comparative Models}\label{sec:methods:models}

We select the two SOTA models UTAE \citep{pastis} and TSViT \citep{tsvit} for comparison. To the best of our knowledge these two models are the closest in terms of architecture and previously used datasets to this work and thus are most suitable for a comparative study. Although both models have seen use on both of our chosen MTLCC and PASTIS datasets, a search indicates that the two models have only ever been used on different tile size variants for PASTIS, and neither have been used on the $48px$ tile size variant of MTLCC. The architectures of these two models are described in section \ref{sec:background:newModels}, with TSViT acting as a comparison for ViT-based models which require spatial patching of tile data, and UTAE as a spatially convolutional model that only uses attention mechanisms for temporal aspects.

There are results for TSViT on MTLCC's $24\times 24$ pixel tile size variant with the 2016 data, but only with the fold index 0 instead of an average across the provided 10 folds, and the results were reported only for the validation set rather than the provided test set. In the case of PASTIS, due to restrictions of available hardware, the $24\times 24$ tile size variant was introduced alongside TSViT. All 5 official PASTIS folds were then used and results were reported for the test sets. Across both datasets, TSViT reports results for a patch size of $2\times2$ pixels.

On the other hand, UTAE was published alongside the PASTIS dataset, and so has prior use on the original $128\times128$ pixel tile size variant, with all 5 official folds used and the test set mean results reported. UTAE does not report results for MTLCC in the original publication, but does have prior use on the same MTLCC $24\times24$ pixel tile size variant and conditions as TSViT through part of TSViT's comparative model study. In contrast, it appears that UTAE and other comparison models in that study remained on the $128\times128$ pixel PASTIS tiles instead of the $24\times 24$ pixel variant. We therefore believe our study is the first to compare our chosen models in the same conditions across all 4 dataset variants.

The UTAE and TSViT authors comprehensively rule out other architectures as SOTA competitors \citep{pastis,tsvit}, therefore we do not include additional past models in our main comparison against the two most successful current methods. We do include UNet3D \citep{unet3d} as a baseline convolutional model with neither attention mechanisms or use of temporal band-date metadata, which our model and the other comparisons all use. All comparison models are run with their original hyperparameters within our PyTorch training, validation, and test framework (Section \ref{sec:methods:training}).

\subsection{Segmentation Metrics}\label{sec:methods:metrics}

In our experiments, we calculate the micro Overall Accuracy (OA\%) and the macro mean Intersection over Union (mIoU). OA\% provides a measure of the total number of correct versus incorrect pixels across the results, while mIoU gives a class-averaged measure of the overlap between ground truth and predicted label masks. Between the two, mIoU is the preferred metric for true segmentation performance, because it is sensitive to both strong or poor performance on minority classes that may be obscured by the pixel-wise (OA\%) accuracy metric \citep{wang2023metrics}. Furthermore, as we have imbalanced data, micro-averaging of OA\% will lead to large classes such as backgrounds or meadows giving an overweight contribution to results \citep{KOCAKmetrics}. Despite these issues, we retain it for direct comparability with other studies on our datasets \citep{pastis,tsvit,mtlcc}. In our main results we also report the Dice Similarity Coefficient (known as F1 Score), which is the harmonic mean of precision and recall, to broaden the set of measures for future comparisons and provide an additional view of model performance under large variations in parcel size and class balance \citep{RTASNetSegmentation,KOCAKmetrics}. Although often used, we do not report the Kappa statistic \citep{kappa}, as it is not comparable across datasets with different numbers of classes \citep{medicalsegmentationmetrics}.

None of these metrics explicitly provide a measure of inter-class boundaries, which we believe to be crucial for SITS crop segmentation, as agricultural parcels often share boundaries but differ in the crops grown on either side \citep{nzCropBoundariesMethods}. A model may accurately predict the crop within a parcel but still leak those predictions into background pixels or neighbouring parcels. A boundary quality metric would measure true model performances in this problem space. Therefore, we selected Boundary IoU \citep{cheng2021boundary} to improve our analysis. The authors believe this is the first use of Boundary IoU (Equation \ref{eq:BIOU}) in SITS crop segmentation.

\begin{align}\label{eq:BIOU}
\text{Boundary IoU}(G, P) = \frac{|(G_d \cap G) \cap (P_d \cap P)|}{|(G_d \cap G) \cup (P_d \cap P)|},
\end{align}
Equation \ref{eq:BIOU} defines the original binary class metric, where $G$ and $P$ are the ground truth and predicted masks for a given class, $d$ is a pixel-distance parameter, and $G_d$ and $P_d$ are the masks restricted to pixels within distance $d$ of their boundaries. We then adapt this for a multi-class setting with mean Boundary IoU (mBIoU) as shown in Equation \ref{eq:biouAdapted}.
\begin{align}\label{eq:biouAdapted}
\operatorname{mBIoU}=\frac{1}{C}\sum_{c=1}^{C}\frac{|(G^c_d \cap G^c) \cap (P^c_d \cap P^c)|}{|(G^c_d \cap G^c) \cup (P^c_d \cap P^c)|},
\end{align}
where $c$ is a single crop class and $C$ is the set of all crop classes that appear in either the label or predicted classes of a tile. In this work, $d$ is set to $0.05$, i.e., 5\% of the tile’s diagonal, rounded to the nearest pixel. This follows the recommendation of the metric's authors, and corresponds to $2$ pixels for a $24\times 24$ pixel tile, $3$ pixels for a $48\times48$, and $9$ pixels for a $128 \times 128$ pixel tile. Consequently, mBIoU scores are not directly comparable across different tile sizes (as discussed further in section \ref{sec:disc:boundaryiou}).

\subsection{Training and Experimental Setup}\label{sec:methods:training}

We used the Adam optimiser \citep{kingma2014adam} with a weight decay of $1.0e-03$. Models would train for up to 150 epochs, with early stopping allowed after epoch 100 if 20 epochs had been observed with a decline in results from the current best epoch.A learning rate of 5.0e-03 was used, with a cosine annealing scheduler \citep{loshchilov2017sgdr} using 10 warm-up epochs and decaying the learning rate to 2.5e-06 by epoch 150. No warm restarts were used, and a dropout of 0.1 was applied. All models used a batch size of 4, since larger dataset variants could not be trained with higher batch sizes and changing batch sizes would introduce variance and bias.

We used standard cross entropy loss, applying masking only to the `Void' parcel class of the PASTIS dataset discussed in section \ref{sec:methods:data}. All other background pixels were retained for this study, as the ability of models to both discriminate crop-to-crop and crop-versus-non-crop is equally important. This also highlighted the value of including Boundary IoU as an evaluation metric, since a masked background class would remove many of the parcel borders.

Models were trained on three environments: a desktop with an Nvidia RTX 4080 GPU and Intel i9-13900K CPU, a server with an Nvidia RTX A6000 GPU and Intel i9-14900K CPU, and Nvidia A100 nodes on the Supercomputing Wales HPC service (\citeauthor{SCW}). Reported inference times are for the desktop hardware.

\section{Results}\label{sec:results}

\subsection{Transformer Architecture Design Ablation}\label{sec:results:architecture}

\begin{table*}[ht!]
\centering
{\begin{tabular}{@{}ccccccccc@{}}
\toprule
& & & \multicolumn{3}{c}{MTLCC-48} & \multicolumn{3}{c}{PASTIS-128} \\
\cmidrule(lr){4-6} \cmidrule(lr){7-9}
\makecell{Transformers\\$temp$ / $spec$ / $spat$} &
\makecell{Cross\\Attention} &
\makecell{Refine\\Head} &
OA\% & mIoU & mBIoU & OA\% & mIoU & mBIoU \\
\midrule
\checkmark / \checkmark / \checkmark & $S{\leftarrow}T$ & \checkmark & 92.0 & \textbf{0.774} & 0.527 & \textbf{83.3} & \textbf{0.649} & 0.435 \\
\checkmark / \checkmark / \checkmark & $S{\leftarrow}T$ & --         & 91.7 & 0.762 & 0.513 & 82.8 & 0.641 & 0.428 \\
\checkmark / \checkmark / \checkmark & $T{\leftarrow}S$ & \checkmark & 91.6 & 0.756 & 0.509 & 81.9 & 0.612 & 0.394 \\
\checkmark / \checkmark / \checkmark & $T{\leftarrow}S$ & --         & 91.0 & 0.724 & 0.491 & 81.3 & 0.600 & 0.386 \\
\checkmark / \checkmark / \checkmark & $Bi$             & \checkmark & \textbf{92.1} & \textbf{0.774} & \textbf{0.537} & 83.0 & 0.640 & 0.429 \\
\checkmark / \checkmark / \checkmark & $Bi$             & --         & 91.6 & 0.761 & 0.518 & 82.5 & 0.629 & 0.418 \\
\checkmark / \checkmark / \checkmark & --               & \checkmark & 92.0 & 0.773 & \textbf{0.537} & 83.1 & 0.643 & 0.428 \\
\checkmark / \checkmark / \checkmark & --               & --         & 91.6 & 0.764 & 0.524 & 82.5 & 0.634 & 0.416 \\
\cmidrule(lr){1-3} \cmidrule(lr){4-6} \cmidrule(lr){7-9}
\checkmark / \checkmark / --         & $S{\leftarrow}T$ & \checkmark & \textbf{92.1} & 0.770 & 0.495 & \textbf{83.3} & 0.647 & 0.401 \\
\checkmark / \checkmark / --         & $S{\leftarrow}T$ & --         & 91.5 & 0.744 & 0.438 & 82.3 & 0.623 & 0.355 \\
\checkmark / \checkmark / --         & $T{\leftarrow}S$ & \checkmark & 91.6 & 0.753 & 0.474 & 81.2 & 0.587 & 0.350 \\
\checkmark / \checkmark / --         & $T{\leftarrow}S$ & --         & 90.7 & 0.695 & 0.412 & 80.4 & 0.566 & 0.310 \\
\checkmark / \checkmark / --         & $Bi$             & \checkmark & 91.9 & 0.773 & 0.501 & 82.9 & 0.636 & 0.390 \\
\checkmark / \checkmark / --         & $Bi$             & --         & 91.4 & 0.746 & 0.433 & 81.7 & 0.611 & 0.346 \\
\checkmark / \checkmark / --         & --               & \checkmark & 91.9 & 0.765 & 0.492 & 83.0 & 0.639 & 0.397 \\
\checkmark / \checkmark / --         & --               & --         & 91.5 & 0.744 & 0.435 & 82.1 & 0.620 & 0.353 \\
\cmidrule(lr){1-3} \cmidrule(lr){4-6} \cmidrule(lr){7-9}
\checkmark / -- / \checkmark & & \checkmark & 92.0 & 0.772 & 0.533 & 83.0 & 0.643 & \textbf{0.437} \\
\checkmark / -- / \checkmark & & --         & 91.7 & 0.755 & 0.523 & 82.6 & 0.636 & 0.429 \\
-- / \checkmark / \checkmark & & \checkmark & 87.8 & 0.600 & 0.395 & 77.7 & 0.503 & 0.307 \\
-- / \checkmark / \checkmark & & --         & 87.8 & 0.599 & 0.396 & 77.7 & 0.496 & 0.307 \\
\checkmark / -- / --         & & \checkmark & 92.0 & 0.771 & 0.495 & 82.8 & 0.638 & 0.395 \\
\checkmark / -- / --         & & --         & 91.1 & 0.738 & 0.431 & 82.0 & 0.616 & 0.346 \\
-- / \checkmark / --         & & \checkmark & 87.4 & 0.579 & 0.348 & 77.4 & 0.494 & 0.283 \\
-- / \checkmark / --         & & --         & 86.9 & 0.552 & 0.307 & 76.9 & 0.474 & 0.258 \\
-- / -- / \checkmark         & & \checkmark & 88.9 & 0.639 & 0.429 & 78.9 & 0.521 & 0.327 \\
-- / -- / \checkmark         & & --         & 89.2 & 0.628 & 0.425 & 78.7 & 0.514 & 0.320 \\
\bottomrule
\end{tabular}}
\caption{\textbf{Our design ablation study on the core components of our PAtteRNS model.} The ablation study was performed on both MTLCC-48 and PASTIS-128's first folds, with the mean of 3 seeded repeats reported. The cross-attention fuse block element is only present where both temporal and spectral transformers are active for the combination of parallel $cls$ token sets prior to the spatial transformer, with {$S{\leftarrow}T$} representing the Spectral Queries Temporal fuse block and so forth as illustrated in Figure \ref{fig:fuseblocks}. Results are reported as overall percentage accuracy, mean Intersection over Union, and mean Boundary Intersection over Union. \textbf{Bold} = best per metric per dataset.}\label{tab:ablationPrimary}
\end{table*}

We first present the architectural ablation study performed for the design of our PAtteRNS model. The first comparison, shown in Table \ref{tab:ablationPrimary}, examines the architectural design of the PAtteRNS model: the inclusion of each dimensional transformer, the fuse block style used to merge parallel spectral and temporal transformers, and whether a convolutional refinement head is used.

Across both datasets, the full complement of temporal, spectral, and spatial transformers together performs best, with temporal attention shown to play the most important single role in the overall performance. We decide to use Spectral Queries Temporal cross-attention in our full model, as this directional attention showed clear dominance over the Temporal Queries Spectral cross-attention, while also producing especially good performance on PASTIS-128. While Bi-Directional cross-attention also produces strong results on MTLCC-48, we note that performance is not vastly greater than with no cross-attention, and likewise appears to negatively impact performance on PASTIS-128. The more neutral effect of Spectral Queries Temporal compared to Bi-Directional or no cross-attention on MTLCC-48 may indicate that the spectral and temporal patterns contribute more equally on this dataset, while in PASTIS-128 the temporal patterns take a majority share compared to its spectral information. This would be backed up by PASTIS-128 having both longer temporal sequences and fewer spectral bands compared to MTLCC-48. We include all fuse block types in our published model to suit any future use on other datasets that may react differently to ours.

Upon introducing the CRH to models with a parallel temporal and spectral transformer, but without a spatial transformer, a mean improvement of $+$0.8 OA\%,  4.1\% in mIoU, and 13.5\% in mBIoU is observed, marking a very significant impact. When introducing the CRH to the models with all transformers, there is a mean improvement of $+$0.5 OA\%, 1.9\% in mIoU and 2.7\% in mBIoU, demonstrating its wide applicability as well as a remaining advantage in spatial representational capacity over attention mechanisms. Notably, the inclusion of the spatial transformer still brings performance improvements in comparison to the temporal and spectral transformers operating solely as a parallel pair, showing that the CRH is best included as a complement to the spatial transformer, not as a full replacement. This is illustrated by the 13.3\% mBIoU increase for MTLCC-48 and 15.3\% for PASTIS-128 when adding the spatial transformer to the temporal and spectral transformers.

The inclusion of the spectral transformer generally improves segmentation performance, although the magnitude of the improvement is relatively modest. Comparing the best-performing configurations, MTLCC-48 improves from 92.0\% to 92.1\% OA, from 0.772 to 0.774 mIoU, and from 0.533 to 0.537 mBIoU when the spectral transformer is included. A similar trend is observed for PASTIS-128, where OA increases from 83.0\% to 83.3\% and mIoU from 0.643 to 0.649, while mBIoU decreases slightly from 0.437 to 0.435. The strongest PASTIS performance is obtained when spectral-to-temporal cross-attention is used, suggesting that interaction between spectral and temporal representations can further enhance the benefit of spectral modelling. Overall, these results indicate that the spectral transformer provides some additional discriminative information in both datasets, particularly for overall and region-level segmentation performance, although its contribution is smaller than that of the temporal transformer.

Our next ablation examines several adjustable hyperparameters in our model. We also report inference time and parameter count, since all three hyperparameters trade off representational capacity against computational cost. The goal is to reduce model complexity where larger hyperparameters do not improve performance, while identifying those that do and evaluating them individually against their cost. The results of this experiment (Table \ref{tab:ablationHyperparameters}) relate to a baseline model of all three transformers, a simple concatenation fuse block, and the inclusion of the CRH.

\begin{table}[ht!]
\setlength{\tabcolsep}{2.8pt}
\begin{tabular}{@{}cccccccc@{}}
\toprule
\makecell{Patch\\$p$} &
\makecell{Dim\\$d$} &
\makecell{$cls$\\tokens} &
OA\% & mIoU & mBIoU &
\makecell{Inference\\Time (ms)} &
\makecell{Params\\(M)} \\
\midrule
4 & 256 & $k$ & \textbf{92.3} & \textbf{0.779} & 0.533 & 22.95 & 14.94 \\
4 & 128 & $k$ & 92.0 & 0.773 & 0.537 & 12.72 & 4.56 \\
4 & 256 & 1   & 92.1 & 0.773 & 0.535 & 17.01 & 12.70 \\
4 & 128 & 1   & 92.0 & 0.774 & 0.538 & 9.88 & 4.00 \\
\midrule
2 & 256 & $k$ & \textbf{92.3} & 0.774 & 0.564 & 86.66 & 12.87 \\
2 & 128 & $k$ & 92.2 & 0.774 & 0.563 & 37.05 & 3.52 \\
2 & 256 & 1   & 92.2 & 0.773 & 0.564 & 57.19 & 10.63 \\
2 & 128 & 1   & 92.2 & 0.777 & \textbf{0.574} & 26.31 & 2.96 \\
\bottomrule
\end{tabular}
\caption{\textbf{Ablation over patch size $p$, hidden dimension $d$, and number of
$cls$ tokens, reporting the mean of three seeded repeats on just MTLCC-48's first fold.} Inference Time is reported for the mean time per batch of 4 samples over the fold's test set for all three seeds. Other results are reported as overall percentage accuracy, mean Intersection over Union, and mean Boundary Intersection over Union, and model trainable Parameters. \textbf{Bold} = best per metric.}\label{tab:ablationHyperparameters}
\end{table}
\FloatBarrier

Most notably, reducing the patch size $p$ significantly improved boundary reasoning, increasing mBIoU by 5.7\% when using $p=2$ instead of $p=4$. However, other metrics showed little variation, while inference time and model size increased significantly. We therefore consider patch sizes $p=2$ and $p=4$ separately for the relevant models. The size of the $cls$ token representation and the embedded dimension size $d$ both showed low variance in performance metrics. Increasing the embedding size from 128 to 256 more than triples model parameters and roughly doubles inference time. These costs were considered too high for the small performance gain, so we kept $d=128$ for the full model. For the $cls$ token, the added inference time and parameters are marginal, so we keep its representation size at $k$ for comparability with TSViT.

Our ablation study also examined two more factors, the explicit masking of temporally padded values in attention, and the choice of date encoding method. Table \ref{tab:ablationExtras} reports these results. In the case of date encoding methods, switching between DOY and DSRD encodings had no notable effect on either dataset. This is unsurprising for MTLCC, whose first capture date lies near the start of the year, but PASTIS spans September 2018 to October 2019 and was expected to be more sensitive; nonetheless, we observed little variation. Given this inconclusive evidence, we adopt DSRD encoding and explicit pad-value masking in the final model to standardise potential inter-dataset differences in temporal window position and completeness. The factor of masking padded temporal bands in attention similarly produced minimal change, suggesting the architecture can largely self-regulate attention to static, unique pad values. We still retain explicit masking of padded tokens in the experiments of section \ref{sec:results:modelcomparisons} to avoid any unforeseen variance due to these pad values.

\def\arraystretch{1.2}
\begin{table}[ht!]
\small
\setlength{\tabcolsep}{3pt}
\begin{tabular}{@{}cccccccc@{}}
\toprule
& & \multicolumn{3}{c}{MTLCC-48} & \multicolumn{3}{c}{PASTIS-128} \\
\cmidrule(lr){3-5} \cmidrule(lr){6-8}
\begin{tabular}[c]{@{}c@{}}Pad\\Mask\end{tabular} &
\begin{tabular}[c]{@{}c@{}}Date Encoding\\Method\end{tabular} &
OA\% & mIoU & mBIoU &
OA\% & mIoU & mBIoU \\
\midrule
Y & DSRD & 91.9 & 0.771 & 0.532 & 82.9 & 0.638 & 0.424 \\
Y & DOY  & 92.1 & 0.775 & 0.530 & 82.9 & 0.637 & 0.423 \\
N & DSRD & 92.0 & 0.773 & 0.537 & 83.1 & 0.643 & 0.428 \\
N & DOY  & 91.9 & 0.772 & 0.526 & 82.9 & 0.640 & 0.424 \\
\bottomrule
\end{tabular}
\caption{\textbf{Ablation over padding mask and temporal date encoding scheme, evaluated on both MTLCC-48 and PASTIS-128's first folds.} Results are reported as overall percentage accuracy, mean Intersection over Union, and mean Boundary Intersection over Union.}\label{tab:ablationExtras}
\end{table}
\def\arraystretch{1}

\subsection{Model Performance Comparisons}\label{sec:results:modelcomparisons}

The results of our comparative experiments are shown in Table \ref{tab:mainsizevariants}. These required substantial computation to achieve a sufficiently deep set of repeats to validate the impact of changing tile size in each dataset, using three differently seeded repeats of every fold on both datasets. All model–dataset combinations produced results, except TSViT with patch size $p=2$, which exceeded GPU memory on all available hardware.

\begin{table*}[ht!]
\begin{threeparttable}
\begin{tabular*}{\textwidth}{@{\extracolsep\fill}llcccc@{}}
\toprule
Model & Patch Size\tnote{i} & OA\% & mIoU & Dice & mBIoU\tnote{ii} \\
\midrule
\textbf{MTLCC 2016} & & \multicolumn{4}{c}{$48\times 48$ pixel tiles / $24\times 24$ pixel tiles} \\
\cmidrule(lr){1-1}\cmidrule(lr){3-6}
UNet3D & -- & 90.3/90.0 & 0.694/0.726 & 0.799/0.826 & 0.448/0.463 \\
\addlinespace
UTAE   & -- & \textbf{92.6}/92.1 & 0.782/0.790 & 0.867/0.874 & 0.550/0.542 \\
\addlinespace
\multirow{2}{*}{TSViT}
& 4 & 91.0/90.4 & 0.749/0.755 & 0.844/0.849 & 0.538/0.510 \\
& 2 & 91.5/90.9 & 0.757/0.769 & 0.849/0.859 & 0.566/0.554 \\
\addlinespace
\multirow{2}{*}{\makecell{PAtteRNS\\ \textit{(proposed)}}}
& 4 & 92.2/91.5 & 0.783/0.789 & 0.869/0.874 & 0.546/0.520 \\
& 2 & 92.4/91.9 & 0.786/\textbf{0.801} & 0.870/\textbf{0.882} & \textbf{0.589}/\textbf{0.576} \\
\midrule
\textbf{PASTIS} & & \multicolumn{4}{c}{$128\times128$ pixel tiles / $24\times 24$ pixel tiles}\\
\cmidrule(lr){1-1}\cmidrule(lr){3-6}
UNet3D & -- & 82.0/82.3 & 0.604/0.618 & 0.732/0.745 & 0.376/0.317 \\
\addlinespace
UTAE   & -- & \textbf{83.8}/83.7 & 0.650/0.638 & 0.773/0.761 & 0.427/0.340 \\
\addlinespace
\multirow{2}{*}{TSViT}
& 4 & 82.3/82.2 & 0.628/0.624 & 0.755/0.752 & 0.444/0.344 \\
& 2\tnote{iii} & --/82.4 & --/0.626 & --/0.753 & --/0.366 \\
\addlinespace
\multirow{2}{*}{\makecell{PAtteRNS\\ \textit{(proposed)}}}
& 4 & 83.5/83.3 & 0.650/0.654 & 0.772/0.777 & \textbf{0.445}/0.357 \\
& 2 & 83.6/83.7 & 0.646/\textbf{0.659} & 0.768/\textbf{0.779} & 0.436/\textbf{0.393} \\
\bottomrule
\end{tabular*}
\begin{tablenotes}\footnotesize
    \item[i] Only ViT-based models utilise the Patch Size parameter
    \item[ii]As stated in section \ref{sec:methods:metrics}, Boundary IoU is only directly comparable within tiles of the same pixel dimensions, and as such has a highlighted top performance for per tile size.
    \item[iii]This combination of dataset, tile-size variant, and patch size was not computationally viable for TSViT.
\end{tablenotes}
\end{threeparttable}
\caption{\textbf{Comparison of state-of-the-art methods in semantic segmentation on dataset tile-size variants.} All reported results are the mean of 3 iteratively-seeded repeats of the mean performance across all folds in a dataset. For each fold, the best epoch by test set performance is considered. All results in these tables are our results, as opposed to results stated in other works. Results are reported as overall percentage accuracy, mean Intersection over Union, Dice Score, and mean Boundary Intersection over Union. \textbf{Bold} = best per metric per dataset, and for mBIoU, per tile-size\textsuperscript{ii}.}\label{tab:mainsizevariants}
\end{table*}

Table \ref{tab:mainsizevariants} shows SOTA results for PAtteRNS. UNet3D, as expected, underperformed the other three models in all cases. Using $p=2$, PAtteRNS holds the highest values for mIoU, Dice, and mBIoU for both datasets by a considerable margin, while UTAE marginally leads in terms of OA\%. Over this metric, these two models notably stand above UNet3D and TSViT. Overall, the results on the MTLCC variants are significantly higher than those achieved on PASTIS. 

\begin{figure*}[ht!]
\centering
\includegraphics[width=0.95\textwidth]{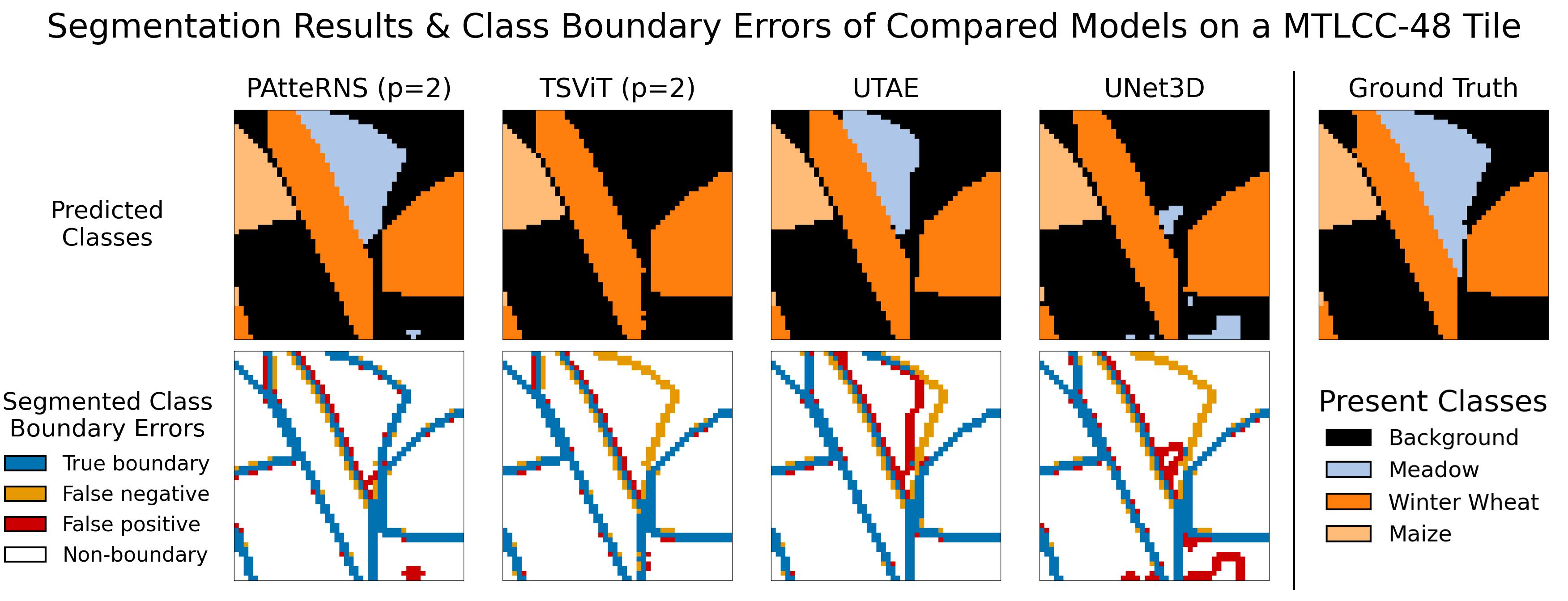}
\caption{\textbf{An illustration of PAtteRNS' segmentation capabilities against comparison models.} Our model is shown here to combine precise boundaries with full recall of present classes in the ground truth labels of a tile from MTLCC-48.}
\label{fig:modelOutputsVisMain}
\end{figure*}

For both ViT-based models, TSViT and PAtteRNS, we generally observe improvements across all metrics when reducing patch size from $p=4$ to $p=2$, in agreement with our initial ablation experiments. This effect is strongest for mBIoU, with average improvements of 8.1\% on MTLCC and 8.2\% on PASTIS-24. However, on PASTIS-128 the TSViT experiment was infeasible due to computational limits, and PAtteRNS showed small decreases in mIoU, Dice, and mBIoU. PAtteRNS largely beats comparison models in mBIoU, as illustrated in Figure \ref{fig:modelOutputsVisMain} and further confirmed by additional results in \ref{app:outputVises}.

Comparing the larger and smaller tile-size variants within each dataset, OA\% is seen to consistently decrease from MTLCC-48 to MTLCC-24 across all models by similar margins, but remains similar from PASTIS-128 to PASTIS-24. In contrast, mIoU and Dice largely improve across both datasets on the smaller tile-size variants, with UNet3D showing particular benefit in its mIoU and Dice increasing by 4.6\% and 3.4\% respectively from MTLCC-48 to MTLCC-24, and 2.3\% and 1.8\% respectively from PASTIS-128 to PASTIS-24. PAtteRNS also shows an increase of 1.9\% and 1.4\% respectively in mIoU and Dice from MTLCC-48 to MTLCC-24, and 2.0\% and 1.4\% respectively from PASTIS-128 to PASTIS-24. There are some standout performance decreases from decreasing tile size, with UTAE showing a decline across all three of OA\%, mIoU, and Dice on PASITS, whilst the same model retains the usual OA\% decrease and mIoU/Dice increase on MTLCC.

Table \ref{tab:speed} shows that, as expected, our PAtteRNS model increases the parameter count relative to other models by adding a third factorised self-attention block for the spectral dimension. However, this increase is not mirrored in the required multiply-accumulate operations (MACs), and PAtteRNS achieves both competitive inference times and the fastest mean epoch duration during training at a patch size of $p=4$. With a patch size of $p=2$, PAtteRNS still has faster inference and mean epoch durations than TSViT at $p=2$, but is slower than UTAE, which does not use a patch-based vision transformer architecture.

Both UTAE and PAtteRNS with $p=4$ have much higher mean best epochs, approaching the 150-epoch limit, suggesting potential performance gains with more epochs, which could not be tested within this study’s time constraints. Table \ref{tab:speed} also shows the cost of reducing patch size from $p=4$ to $p=2$ for TSViT and PAtteRNS in almost all respects. The only clear benefit is a drop in PAtteRNS’s mean best epoch from 140.3 ($p=4$) to 87.7 ($p=2$), slightly offsetting the longer per-epoch training time. For TSViT, however, the mean best epoch increases from 69.7 to 90.0 under the same change, worsening the impact of longer epoch durations.

\def\arraystretch{1}
\begin{table}
\small
\setlength{\tabcolsep}{3pt}
\begin{tabular}{@{}lccccc@{}}
\toprule
Model & {\makecell{Params\\(M)}}  & {\makecell{MACs\\(G)}}  & {\makecell{Inference\\ Time (ms)}} &  {\makecell{Mean Epoch\\ Duration (s)}}& {\makecell{Mean Best\\ Epoch}}\\
\midrule
UNet3D    &  1.6  & 20.2 & 14.0  & 42.7 & 84.0  \\
UTAE   &  1.1 & 9.8 & 24.3  & 98.0 & 135.7  \\
TSViT $P4$  &  1.7 & 9.4 & 13.4  & 59.0 & 69.7  \\
PAtteRNS $P4$ & 3.4 & 11.5 & 13.5  & 40.3 & 140.3  \\
TSViT $P2$   &  1.7 & 36.8 & 56.1  & 167.0 & 90.0  \\
PAtteRNS $P2$ & 3.3 & 44.5 & 37.9  & 143.7 & 87.7  \\
\bottomrule
\end{tabular}
\caption{\textbf{Model Computational Performance Comparison.} Model trainable parameters, MACs of the forward pass, mean inference time, and mean training time are reported. All reported results are for 3 seeded repeats using MTLCC 48's first fold, with inference time reporting the mean time per batch of 4 samples over the test, mean epoch duration reporting the average combined train and validation duration per epoch using our training setup, and best epoch measured as the average index of the best test set mIoU out of 150 maximum epochs from training. MACs are reported for the dimensions of MTLCC 48.}
\label{tab:speed}
\end{table}
\def\arraystretch{1}

\section{Discussion}\label{sec:disc}

\begin{figure*}[ht]
\centering
\includegraphics[width=0.3\textwidth]{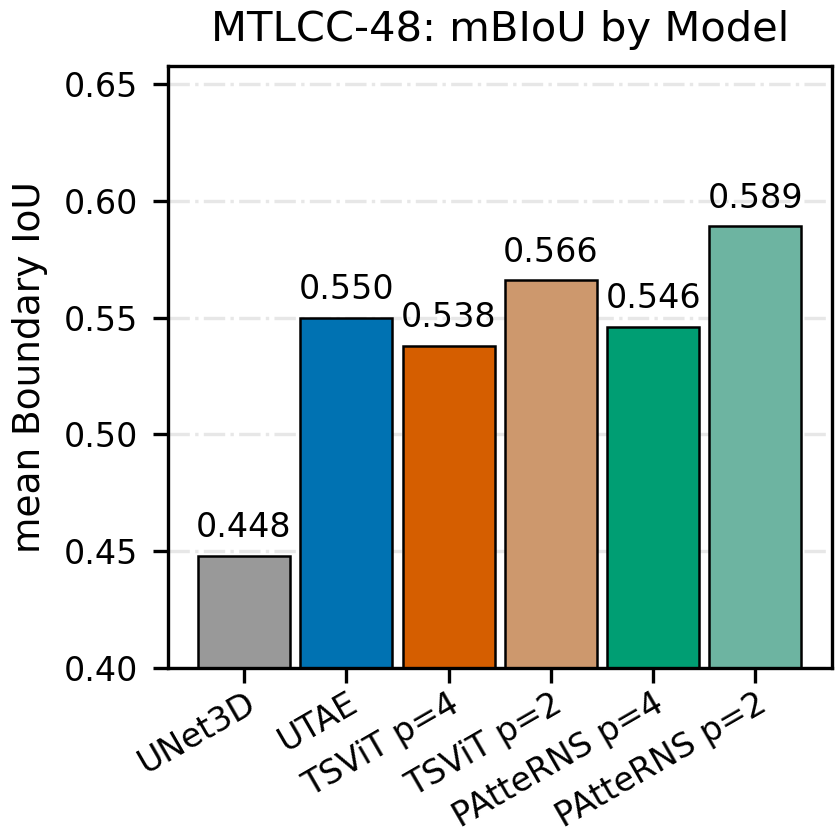}
\hspace{1em}
\includegraphics[width=0.3\textwidth]{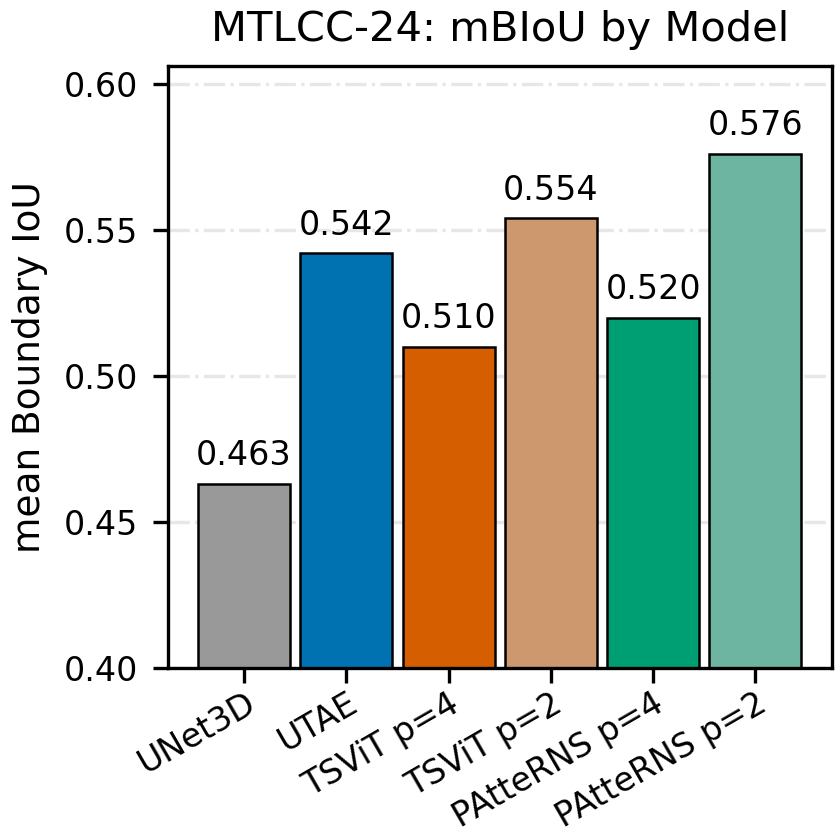}
\\[1em]
\includegraphics[width=0.3\textwidth]{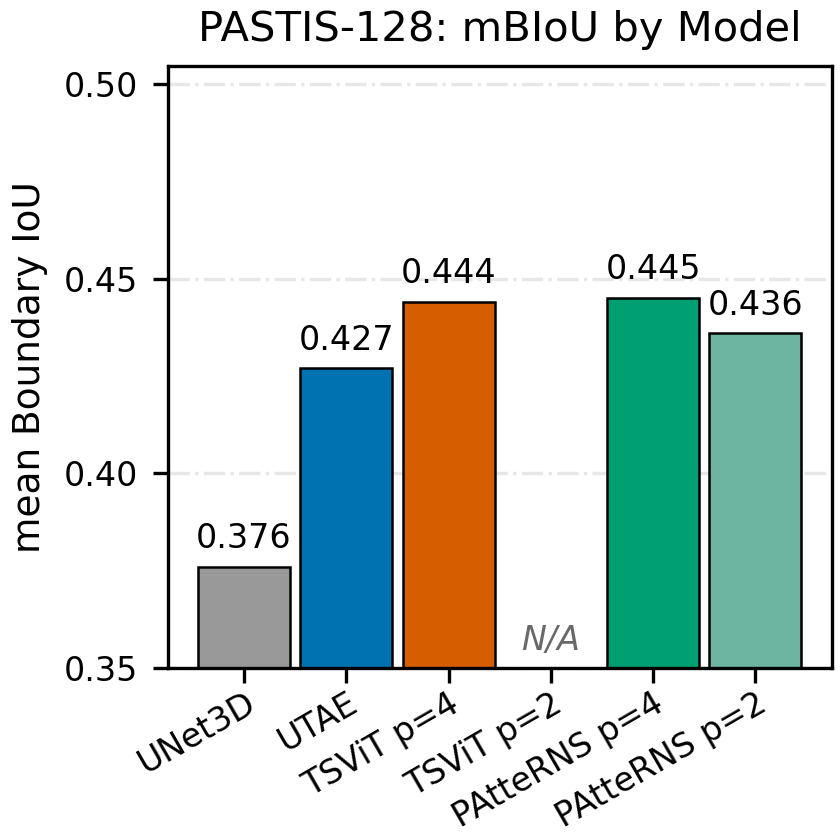}
\hspace{1em}
\includegraphics[width=0.3\textwidth]{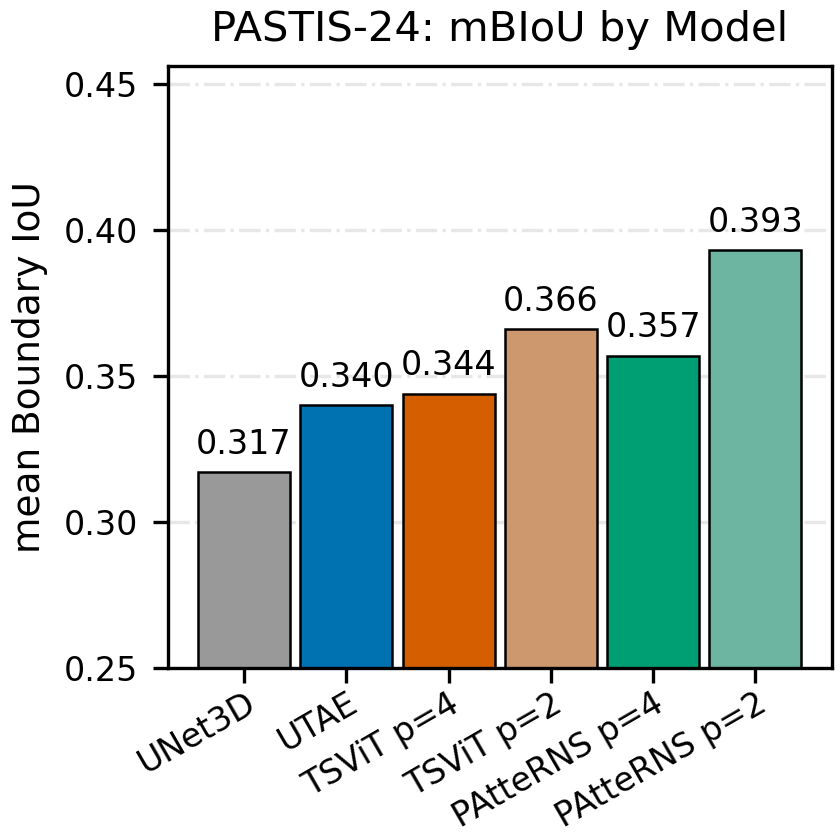}
\caption{\textbf{Mean Boundary IoU (mBIoU) achieved by each model across all four dataset and tile-size combinations.} PAtteRNS with $p=2$ achieves the leading result in three of the four combinations, with PAtteRNS at $p=4$ leading marginally on PASTIS-128 (see Table \ref{tab:mainsizevariants}).}
\label{fig:mbiouGrid}
\end{figure*}

Overall, we achieve the strongest results on nearly all metrics and dataset variants, with the only exception being OA\%, where UTAE leads by at most 0.6\% (mean lead 0.3\%). Despite this, when considering each model's strongest configuration per dataset, an analysis of our model against UTAE still indicates a broader advantage for PAtteRNS. On MTLCC, both models see their best results on MTLCC-24, with PAtteRNS at $p=2$ leading UTAE by 1.4\% in mIoU, 0.9\% in Dice, and 6.3\% in mBIoU, while trailing by -0.2 in OA\%. On PASTIS, UTAE's best results are reported for PASTIS-128, while PAtteRNS's best is achieved with $p=2$ on PASTIS-24. Here, the OA\% gap narrows to just 0.1 percentage points, while PAtteRNS retains a lead of 1.4\% in mIoU and 0.8\% in Dice. As these two configurations differ in tile size, mBIoU cannot be directly compared between them (see section \ref{sec:methods:metrics}), however, for UTAE's strongest configuration on PASITS-128, PAtteRNS at $p=2$ still leads in mBIoU by 2.1\%, with the same comparison on PASTIS-24 raising to a 15.5\% lead. This evidences that under fair comparison, PAtteRNS matches or exceeds UTAE on every metric where a valid comparison can be made.

This conclusion is further supported by the per-class behaviour influencing each model's overall performance in their best configurations (\ref{app:confmatrices}). On MTLCC-24, PAtteRNS achieves a higher true positive rate than UTAE for 13 of 18 classes, while the performance of PAtteRNS at $p=2$ on PASTIS-24 against UTAE on PASTIS-128 shows a lead again for PAtteRNS in 15 of 19 classes, with one class drawn. A further contributor to UTAE's narrow OA\% lead at each dataset's best configuration is the Background class specifically, which represents a substantial share of all pixels in both MTLCC-24 (43.2\%) and PASTIS-24 (44.7\% of non-void pixels). UTAE achieves a higher true positive rate than PAtteRNS on this class in both cases, leading by 1.2 points on MTLCC-24 and 2.6 points on PASTIS-24. Weighted by class population, this single class contributes $+$0.52 and $+$1.16 percentage points to UTAE's OA\% lead on MTLCC-24 and PASTIS-24 respectively, both values exceeding the relevant OA\% lead UTAE holds. These results indicate that UTAE's marginal OA\% advantage is concentrated in the disproportionately large background classes, rather than reflecting any broader advantage in segmentation performance for crop classes across the datasets as a whole.

Table \ref{tab:speed} shows that, with patch size $p=4$, our fully factorised temporal-spectral-spatial attention for SITS crop segmentation matches other SOTA models in both training and inference speed, achieving the shortest mean epoch time and inference within 0.1ms of TSViT and UTAE, implying all three saturate hardware data I/O limits. Our architecture also decouples the parameter cost of the additional factorised spectral self-attention block from its computational cost, increasing representational capacity without a proportional rise in computation.

When reducing patch size from $p=4$ to $p=2$, our model and TSViT generally improve across all metrics (see section \ref{sec:results:modelcomparisons}), further increasing PAtteRNS’s lead over comparison models. However, TSViT still underperforms UTAE in all fields except mBIoU. An exception occurs for our model on PASTIS-128 at $p=2$, where mIoU, Dice, and mBIoU decrease slightly compared to $p=4$, unlike the improvements seen elsewhere. Because this patch–tile combination yields the longest token sequences in our study, this may indicate that the model’s depth and capacity for global self-attention are saturated. Confirming this and exploring whether larger hyperparameters could improve performance will require additional experiments beyond the scope of this work.

For mBIoU, PAtteRNS leads the next-best model by an average of 4.4\% at $p=2$ (both datasets). At $p=4$, it is 2.5\% behind UTAE on both MTLCC variants but still leads by 2.0\% on average across both PASTIS variants. Considerations in section \ref{sec:results:architecture} suggest that retaining the CRH alongside the spatial encoder transformer contributes to this, as their combination enhances fine boundary delineation. The performance in mBIoU of all models is shown across the four dataset tile-size variants in Figure \ref{fig:mbiouGrid}, highlighting the lead PAtteRNS holds across all combinations.

\subsection{Comparison Model Performances}\label{sec:disc:comparisonmodels}

Our results in Table \ref{tab:mainsizevariants} differ from those in the works introducing both UTAE \citep{pastis} and TSViT \citep{tsvit}. For UTAE, higher performance is achieved in this work than the results reported in the original study using the PASTIS-128 dataset, with OA\% increasing from 83.2\% to 83.8\%, and mIoU increasing from $0.631$ to $0.650$. We believe the most likely source of this improvement is the larger number of epochs in our training regime, with the original work only using 100 epochs compared to our 150. This is evidenced in Table \ref{tab:speed}, where UTAE reaches its best-performance at a mean epoch of 135.7, close to the end of the additional 50-epoch training period used in our implementation. Further experiments lie beyond the scope of this study but may be worthwhile for both UTAE and PAtteRNS, which has a mean best epoch of 140.3, to assess performance over many more epochs. For TSViT, we observe the opposite trend: our experiments fail to reproduce the model's originally reported performance on both MTLCC-24 and PASTIS-24. For MTLCC, this appears to be simply due to the original work masking the `Background' class in the loss and metrics, which we do not mask, as discussed in Section \ref{sec:methods:training}. For PASTIS, although the original paper appears to suggest that the `Background' class was masked in a similar manner, our interpretation of the published implementation is that only the `Void' class was excluded from the loss computation and evaluation metrics for PASTIS-24, which matches our implemented masking practice. We could not identify a specific cause for TSViT's performance drop on PASTIS-24 compared to it's original reported results, and whilst UTAE on PASTIS-24 shows a clear performance drop across all comparable metrics relative to PASTIS-128, it's performance on PASTIS-24 in our work still surpasses what was reported as a comparison for TSViT's introduction in that work. TSViT at $p=4$ does improve slightly from PASTIS-24 to PASTIS-128, which makes us conclude that UTAE’s convolutional architecture may require more careful parameter tuning when moving away from its original tile size than the ViT-based TSViT. Based on our experiments, UTAE remains the stronger of these two models for crop segmentation.

\subsection{Boundary IoU}\label{sec:disc:boundaryiou}

Boundary IoU is a useful measure of boundary quality in segmentation results, capturing performance aspects that traditional metrics under-represent. Its main weakness is reliance on a fixed pixel width for the `boundary' region, which makes mBIoU scores incomparable across different tile-size variants, even on the same dataset. Future work could increase the relevance of Boundary IoU for cropland segmentation by defining boundary width in real-world spatial units rather than pixels. Because all SITS data sources have known spatial resolutions, often at standard sizes like 10 m/pixel, this would enable a more scientifically grounded notion of accurate boundary placement and support broader, more consistent adoption of the metric across datasets.

\subsection{Class-wise Performance Across Datasets}\label{sec:disc:perclassissues}

As posed in sections \ref{sec:background:datasets} and \ref{sec:methods:data}, our experiments test whether varying class groupings in crop segmentation datasets reduce reported model performance. We found that the PASTIS-128 dataset's parcels were 60\% non-homogenous in ground coverage, equating to 25.2\% of the pixels in the dataset, while MTLCC-48 sees only 24\% of its parcels featuring classes with non-homogenous ground coverage, comprising of only `Meadow' in this case, measuring in at 14.1\% of dataset pixels. Additionally, four PASTIS classes were considered taxonomical aggregates: `Fruits, Vegetables, \& Flowers`, `Leguminous Fodder`, `Orchard`, and `Mixed Cereal`. These were expected to perform poorly in between-class confusion because they lack a single phenological identity and may biologically overlap with other classes. In MTLCC, `Winter Wheat' was found as a taxonomical aggregate of soft and hard winter wheats, and `Rapeseed' as a seasonal aggregate of its own seasonal varieties, although these classes are both potentially less damaging to between-class confusion than the aggregates found in PASTIS. Beyond the large advantage MTLCC shows over PASTIS for all models and metrics in Table \ref{tab:mainsizevariants}, our analysis (\ref{app:badclassanalysis}) indicates that PASTIS’s avoidable taxonomical aggregate classes, and the taxonomically correct but non-homogenous classes in both datasets substantially reduce model performance.

\section{Conclusion}\label{sec:conclusion}

This study introduces our new model for SITS crop semantic segmentation, PAtteRNS, and shows our novel fully-factorised attention architecture to achieve SOTA results for this field. This performance advantage is achieved at similar computational cost to existing models at patch size $p=4$, with a greater margin obtained at $p=2$ for a higher computational cost. Although this smaller patch size leads to slower training and inference than non-ViT-based models, our model still surpasses TSViT, the incumbent SOTA ViT-based model, for this patch size in speed and capability.

Our model exhibits particularly strong performance in parcel boundary delineation, demonstrated through the Boundary IoU metric, which we show to be a valuable criterion for the quantitative evaluation of crop segmentation models. Application of a convolutional refinement head to spatial transformer outputs was shown as a key part of this strength, demonstrating the continued benefit of convolutional components in contemporary transformer-heavy crop segmentation models.

We also find that crop segmentation models perform differently when trained and evaluated with SITS tiles of different pixel sizes, and therefore different tile sizes should not be used in fair comparison. While many models improve with smaller tile sizes, others loose significant capacity, especially if designed for larger tiles. Our findings here indicate substantial scope for future research on principled methods to determine the optimal tile size for models operating on any given dataset. Such methods could enable dynamic adjustment of tile dimensions during training, followed by reassembly of the tiles for inference and evaluation, and may ultimately yield performance gains for crop segmentation models. Our experiments suggest that smaller tile sizes may work to the advantage of models utilising global self-attention mechanisms.

Finally, we provide evidence that taxonomical aggregate classes are a potential cause of poor crop segmentation performance across all studied models, and suggest that future datasets should work to minimise the number of taxonomical aggregations when constructing class sets from ground truth data sources. We also find that non-homogeneous taxonomic classes tend to lower reported performance, and models must improve at handling these valid and common classes. We could not confirm any effects of seasonal aggregate classes due to a low population of this aggregation type within our chosen datasets, however this should be taken into consideration for future investigation. Further research on true multi-year datasets is also needed to assess the importance of positional encoding methods for capture dates in deep temporal models.

\section*{Author Contributions}

CRediT Roles: \textbf{Joseph Metcalfe} Conceptualisation, Methodology, Data Curation, Investigation, Software, Formal Analysis, Validation, Visualisation, Writing - original draft, Writing - review \& editing. \textbf{Sara Sharifzadeh} Conceptualisation, Supervision, Funding acquisition, Resources, Writing - review \& editing. \textbf{Fabio Caraffini} Supervision, Resources, Writing - review \& editing.

\section*{Acknowledgements}

Funding for this paper was kindly provided by Swansea University EPSRC DTP funding Project Reference EP/W524694/1. The funding source had no involvement or say in the contents of this paper. We also acknowledge the great support of the Supercomputing Wales project, which is part-funded by the European Regional Development Fund (ERDF) via Welsh Government.

\section*{Declaration of Competing Interest}

The authors declare that they have no known competing financial interests or personal relationships that could have appeared to influence the work reported in this paper.

\section*{Data Availability Statement}

All code and models associated with this work are available in the PAtteRNS GitHub repository (\url{https://github.com/JoeMetc/CroplandPAtteRNS}). The first dataset used, PASTIS \citep{pastis}, is available at \url{https://doi.org/10.5281/zenodo.5012942} with $128px$ tiles available natively and $24px$ tiles available through the provided re-size processing code in the PAtteRNS GitHub. The second dataset used, MTLCC \citep{mtlcc}, is available at \url{https://doi.org/10.3390/ijgi7040129} with both $48px$ and $24px$ tiles available natively.

\bibliographystyle{elsarticle-harv} 
\bibliography{references}

\appendix

\clearpage

\onecolumn

\section{Dataset Class Comparison Table}\label{app:datasetClasses}

\begin{table}[htbp!]
\centering
{\begin{threeparttable}
\begin{tabular}{@{}lccl@{}}
\toprule
Class Name & PASTIS ID & MTLCC ID & Aggregation \\
\midrule
Background\tnote{*}                  & 0  & 0  &          \\
Meadow                                & 1  & 3  & NH      \\
Winter Soft Wheat\tnote{*}            & 2  & -- &         \\
Maize\tnote{*}                        & 3  & -- &         \\
Winter Barley                         & 4  & 14 &         \\
Winter Rapeseed                       & 5  & -- &         \\
Spring Barley                         & 6  & -- &         \\
Sunflower                             & 7  & -- &         \\
Grapevine                             & 8  & -- & NH      \\
Sugar Beet\tnote{*}                   & 9  & 1  &         \\
Winter Triticale                      & 10 & 7  &         \\
Winter Durum Wheat                    & 11 & -- &         \\
Fruits, Vegetables, \& Flowers        & 12 & -- & NH, TA  \\
Potatoes\tnote{*}                     & 13 & 10 &         \\
Leguminous Fodder                     & 14 & -- & NH, TA  \\
Soybeans                              & 15 & 11 &         \\
Orchard                               & 16 & -- & NH, TA  \\
Mixed Cereal                          & 17 & -- & NH, TA  \\
Sorghum                               & 18 & -- &         \\
Void                                  & 19 & -- &         \\
Summer Oat                            & -- & 2  &         \\
Rapeseed\tnote{*}                     & -- & 4  & SA      \\
Hop                                   & -- & 5  &         \\
Winter Spelt                          & -- & 6  &         \\
Beans                                 & -- & 8  &         \\
Peas                                  & -- & 9  &         \\
Asparagus                             & -- & 12 &         \\
Winter Wheat                          & -- & 13 & TA      \\
Winter Rye                            & -- & 15 &         \\
Summer Barley                         & -- & 16 &         \\
\bottomrule
\end{tabular}
\begin{tablenotes}\footnotesize
    \item[*] Marked classes have been renamed in one or both datasets to match equivalent classes between datasets or to improve clarity from the original class names.
\end{tablenotes}
\end{threeparttable}}
\caption{\textbf{Correspondence between PASTIS and MTLCC crop classes.} Classes are ordered first by PASTIS ID, then by remaining MTLCC-exclusive class ID. Classes are annotated with the issues and flawed groupings introduced in section \ref{sec:background:datasets}, where NH $=$ Non-Homogeneous, TA $=$ Taxonomical Aggregate, and SA $=$ Seasonal Aggregate.}\label{tab:classes}
\end{table}

\twocolumn

\section{Class-wise Performance Analysis}\label{app:badclassanalysis}

To better understand the impact of class grouping on crop segmentation datasets, we analyse confusion matrices from our three main models (\ref{app:confmatrices}) to observe the potential effects on model overall performance contributed by the three identified themes of classes, `Non-Homogenous', `Taxonomical Aggregates', and `Seasonal Aggregates' (See sections \ref{sec:background:datasets}, \ref{sec:methods:data}, and \ref{sec:disc:perclassissues}). 

For MTLCC, the highest individual error is seen in around half of `Meadow' class pixels being misidentified as `Background' by all models. This is a classic overlap, and one of the hardest boundaries to define outside the ground truth data, but still represents the only non-homogenous class of the dataset making up a significant portion of the drop in overall performance. The taxonomical aggregate class, `Winter Wheat', also provides a significant amount of the error rates through false positives on pixels belonging to the `Winter Spelt' and `Winter Triticale' classes, which is to be expected, given that both spelt and triticale have biological relation to common wheat. The seasonal aggregate class, `Rapeseed', is actually the highest performing class in the dataset for all three models, which could indicate a combination of unique spectral-temporal phenology with the narrow geographical window of MTLCC lending most parcels to being the same seasonal variant.

Within PASTIS, `Meadow' classes are falsely classified as `Background' to a much lower proportion, but this still represents one of the largest errors in the dataset's results due to the class balance of `Meadow' being very strong. The other solely non-homogenous class, `Grapevine', also sees a high rate of misidentification as `Background' by all models. All four of PASTIS' taxonomical aggregate classes perform poorly, with a mean recall across these classes in PASTIS-24 of 60.6\% for PAtteRNS $p=2$, 54.6\% for TSViT $p=2$, and 52.5\% for UTAE. In comparison, the set of classes excluding taxonomical aggregates stands at an average recall of 83.4\%, 81.0\%, and 80.8\% respectively for the same models. `Mixed Cereal', predictably, sees scattered but significant confusion with a number of cereals, likely due to the biological overlap. `Leguminous Fodder' is mostly confused with `Background' and `Meadow', which could be due to a lack of solid internal class phenology. `Orchard' is primarily confused with `Background', backing the theory of class phenological representation influence from the grasses interspacing the fruit trees. Further evidence for this comes from it's second and third most common confusions, `Grapevine' and `Meadows`, both of which have been identified in our review as non-homogenous classes which may have various similar grasses populating their spatial extent. `Fruits, Vegetables, \& Flowers' again has major overlap with `Background', but also sees notable confusion with `Sunflower' and `Potatoes' across all models, very likely attributable to biological overlap in these classes due to the taxonomical aggregation.

The confusion matrices for PASTIS-24 also show poor performance on `Sorghum', which was not one of our flagged classes, but was a relative minority population class, This class saw most confusion with 'Maize', which can be explained by close biological relation between the two. The other consistently weak class we saw in PASTIS was `Winter Triticale', which saw a lot of false classification as `Soft Winter Wheat' and `Mixed Cereal', again bringing the influence of a taxonomical aggregate class into play. Overall, models reported vastly higher performances when trained with MTLCC rather than PASTIS, supporting the theory of taxonomical aggregate classes limiting the performance achievable by models on crop segmentation datasets.

\FloatBarrier

\section{Additional Model Output Visualisations}\label{app:outputVises}

\begin{figure*}[htb]
\centering
\includegraphics[width=0.95\textwidth]{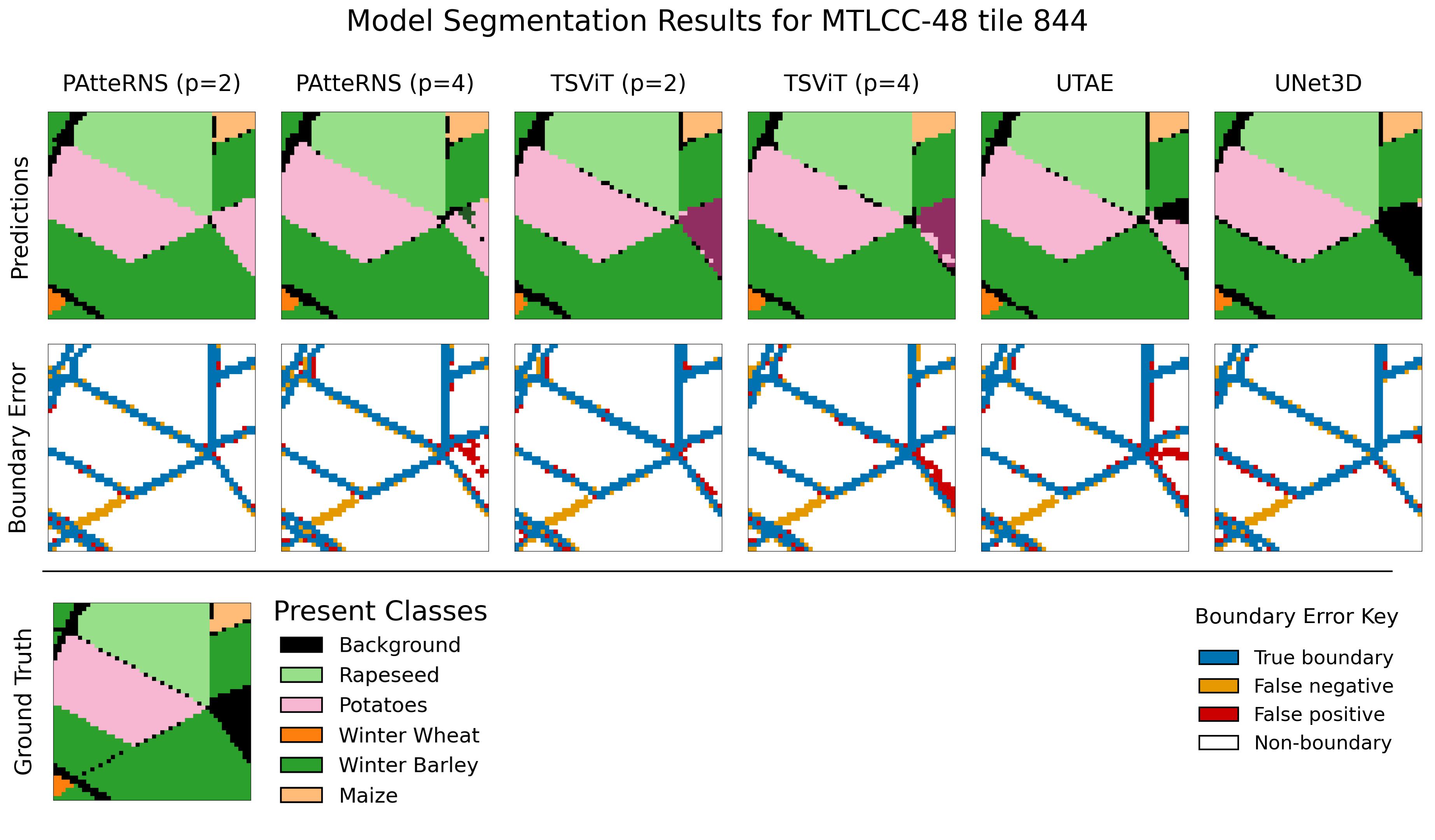}
\caption{Class segmentation outputs, and boundary error maps, for all models on an example tile from MTLCC-48.}
\label{fig:appBmtlcc48b34}
\end{figure*}

\begin{figure*}[htb]
\centering
\includegraphics[width=0.95\textwidth]{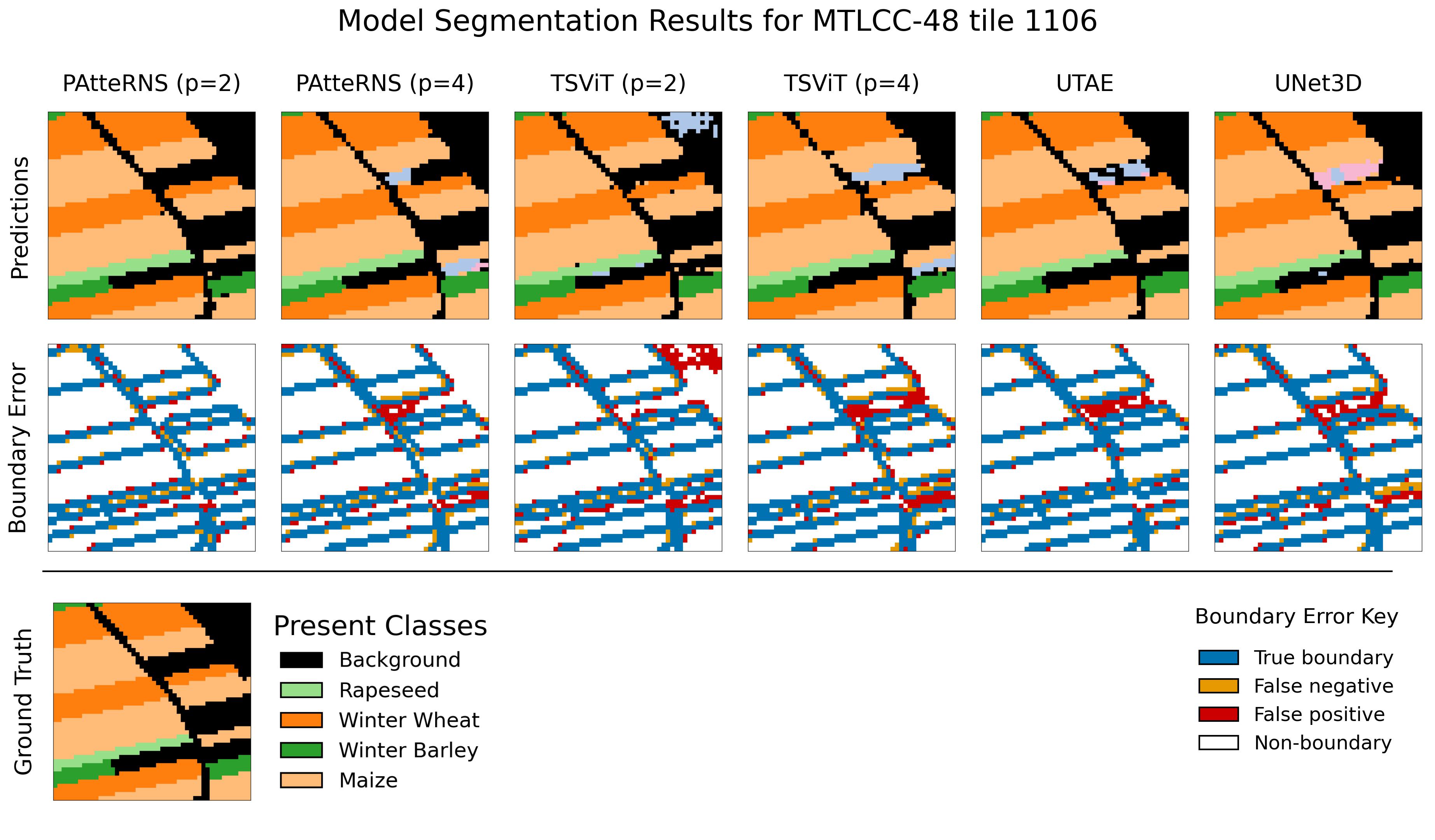}
\caption{Class segmentation outputs, and boundary error maps, for all models on an example tile from MTLCC-48.}
\label{fig:appBmtlcc48b56}
\end{figure*}

\begin{figure*}[htb]
\centering
\includegraphics[width=0.95\textwidth]{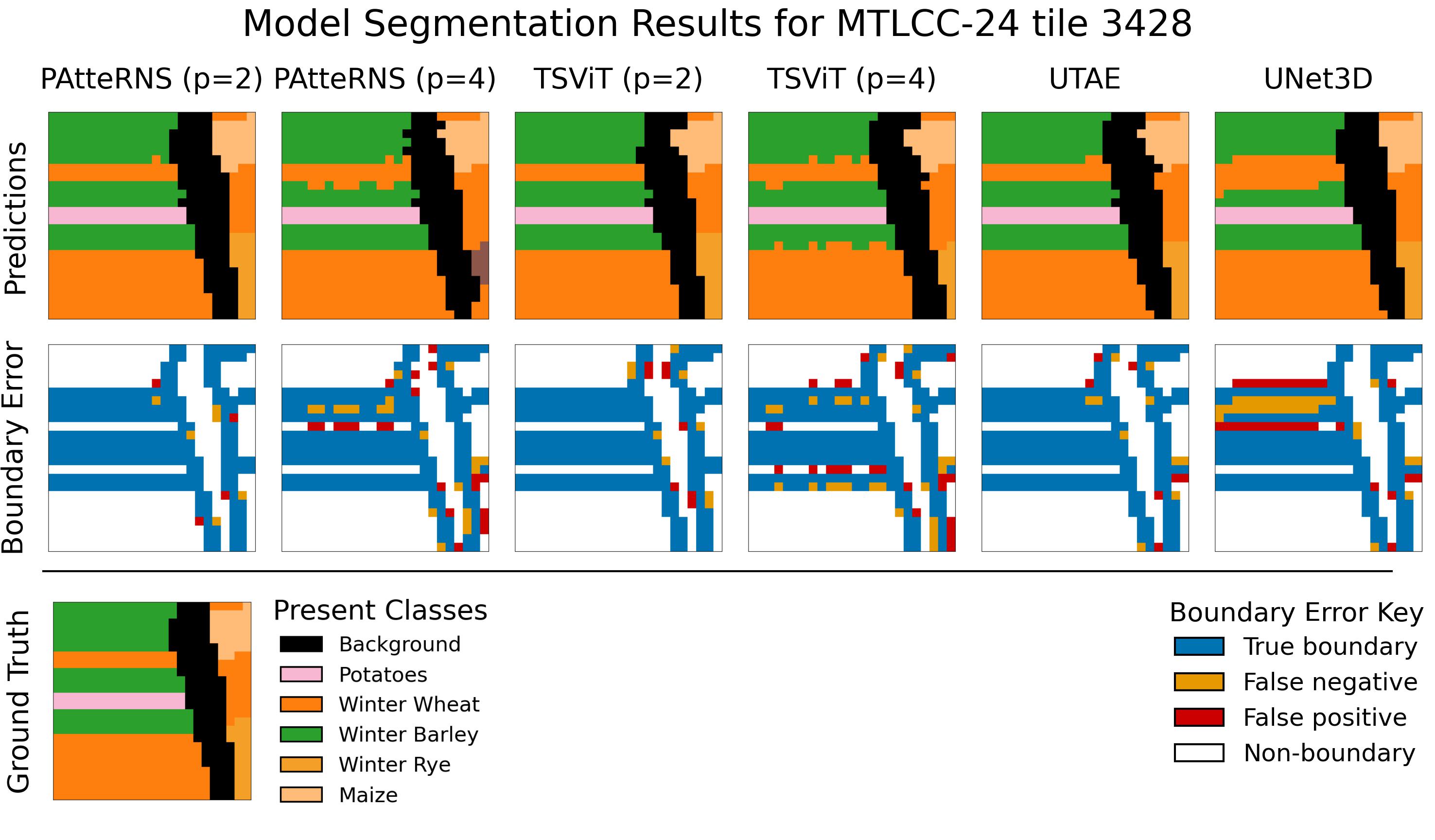}
\caption{Class segmentation outputs, and boundary error maps, for all models on an example tile from MTLCC-24.}
\label{fig:appBmtlcc24b6}
\end{figure*}

\begin{figure*}[htb]
\centering
\includegraphics[width=0.95\textwidth]{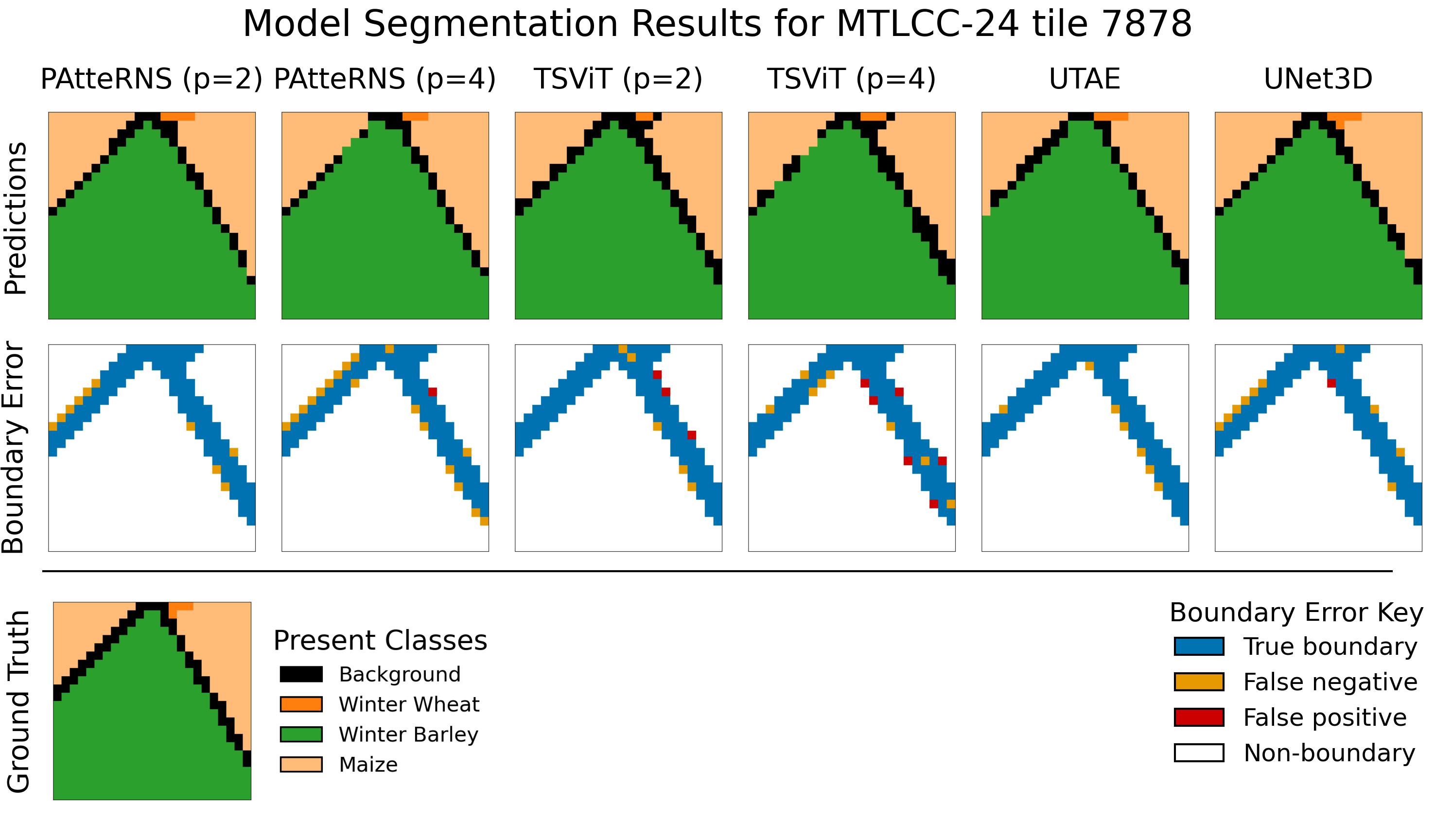}
\caption{Class segmentation outputs, and boundary error maps, for all models on an example tile from MTLCC-24.}
\label{fig:appBmtlcc24b147}
\end{figure*}

\begin{figure*}[htb]
\centering
\includegraphics[width=0.8\textwidth]{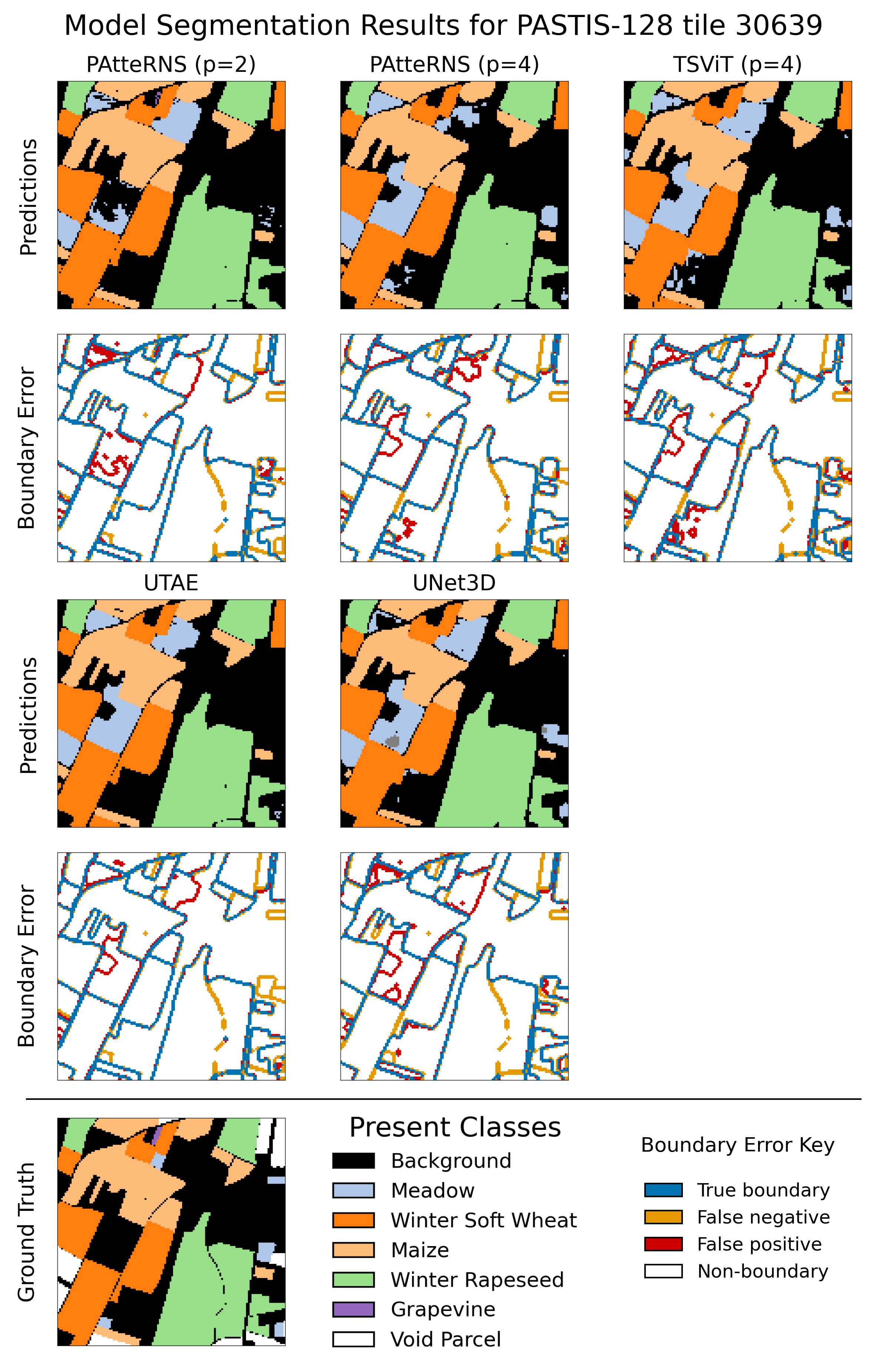}
\caption{Class segmentation outputs, and boundary error maps, for all viable models on an example tile from PASTIS-128.}
\label{fig:appBpastis128b96}
\end{figure*}

\begin{figure*}[htb]
\centering
\includegraphics[width=0.8\textwidth]{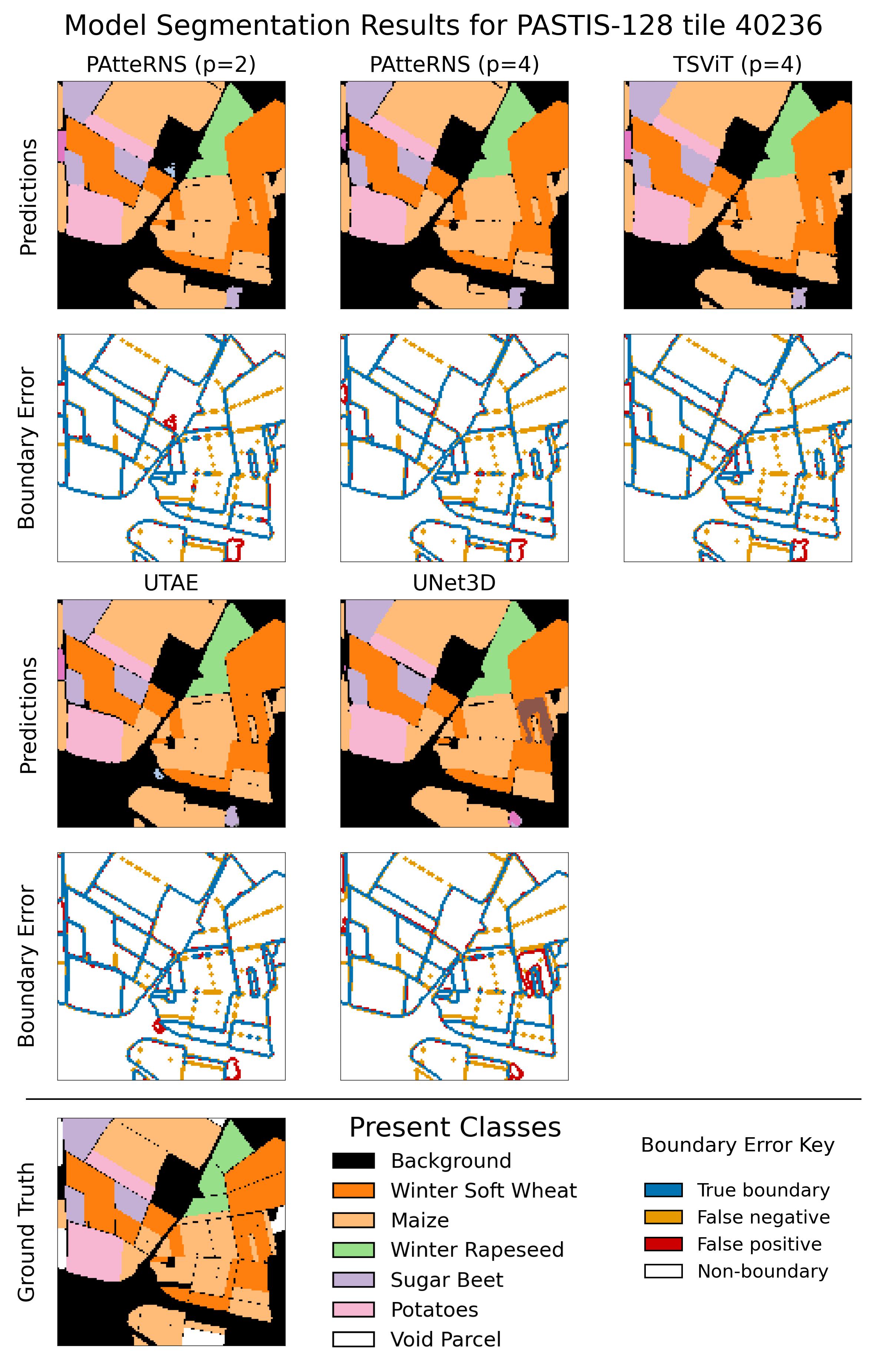}
\caption{Class segmentation outputs, and boundary error maps, for all viable models on an example tile from PASTIS-128.}
\label{fig:appBpastis128b110}
\end{figure*}

\begin{figure*}[htb]
\centering
\includegraphics[width=0.95\textwidth]{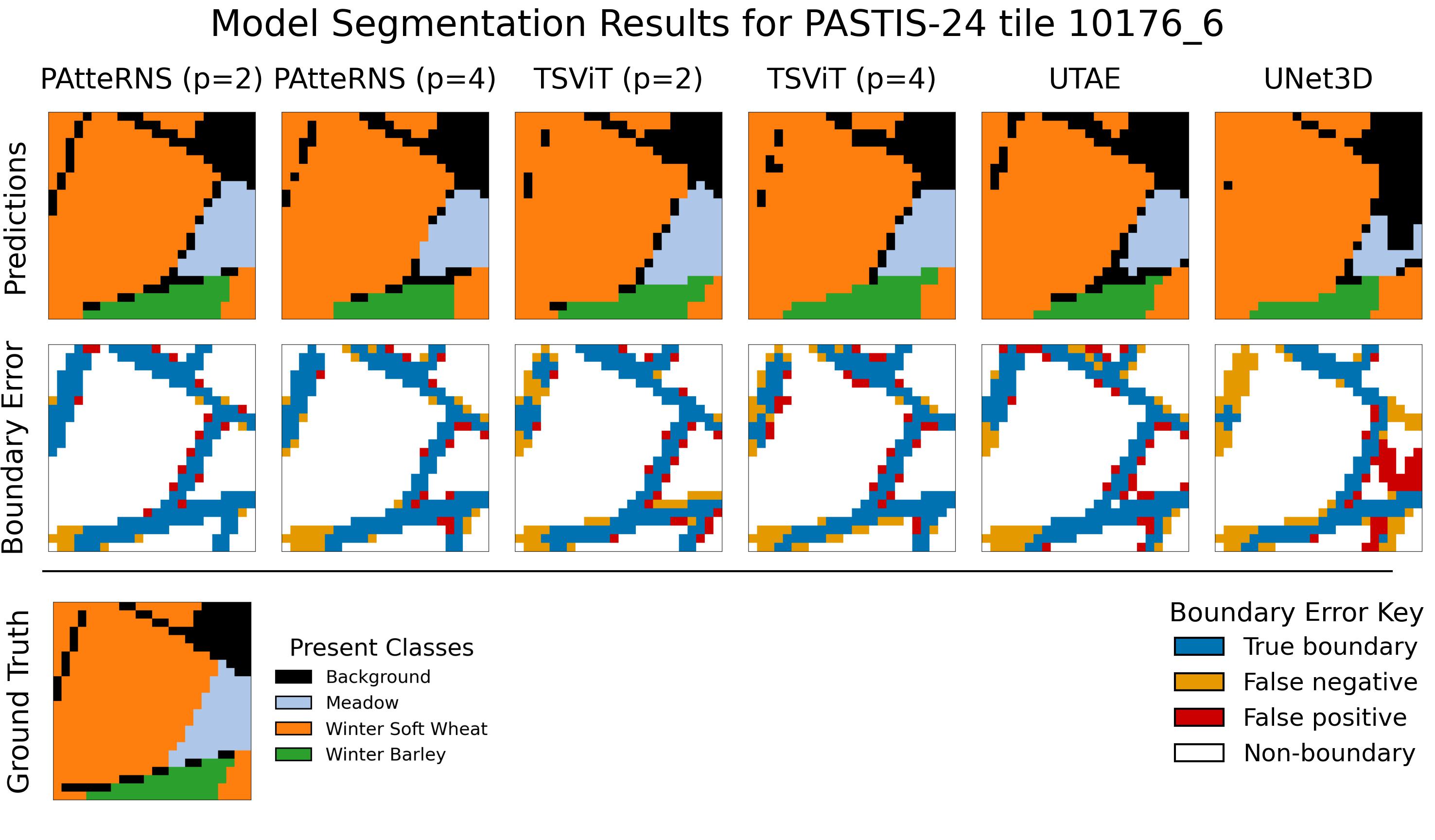}
\caption{Class segmentation outputs, and boundary error maps, for all models on an example tile from PASTIS-24.}
\label{fig:appBpastis24b264}
\end{figure*}

\begin{figure*}[htb]
\centering
\includegraphics[width=0.95\textwidth]{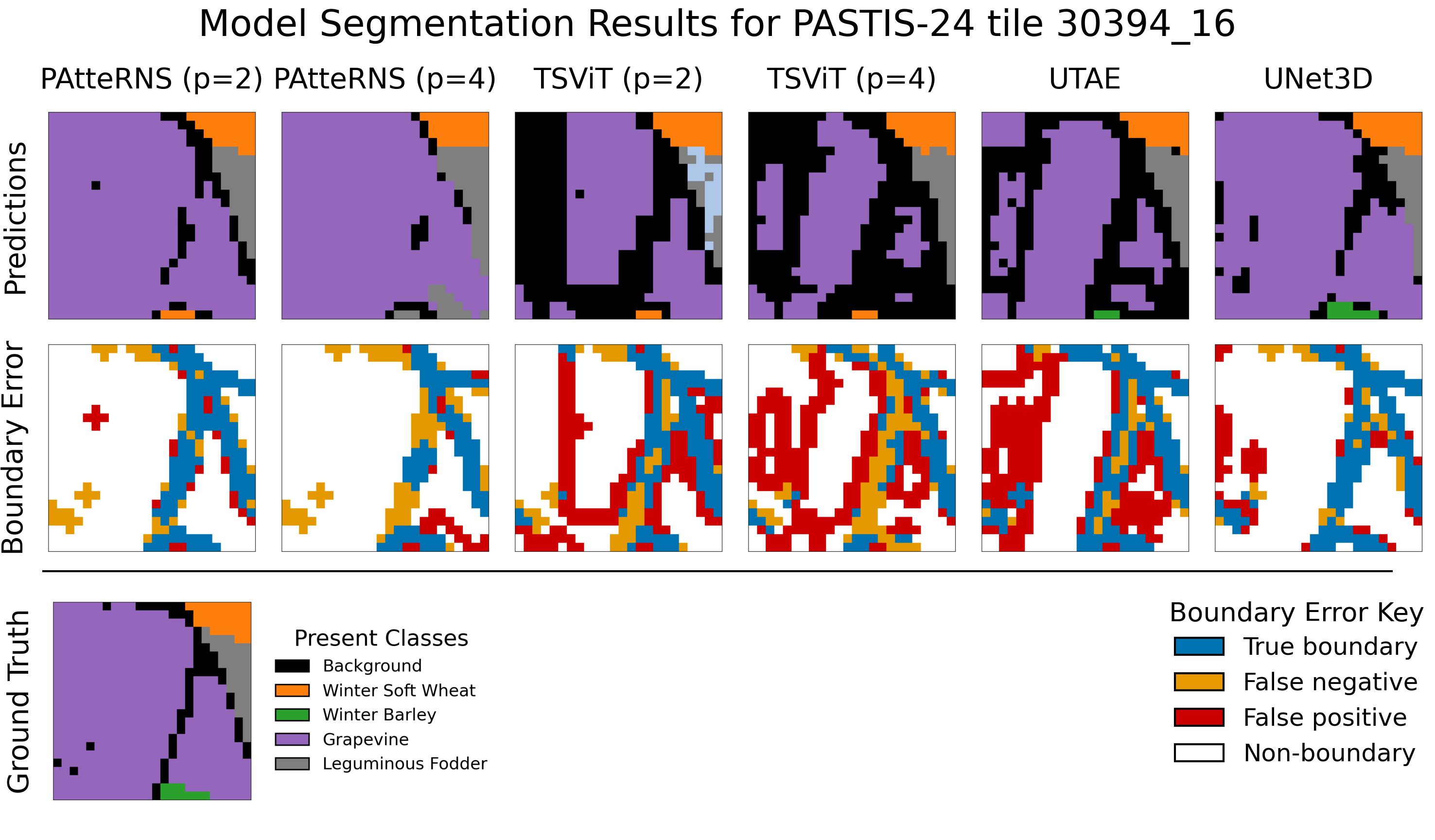}
\caption{Class segmentation outputs, and boundary error maps, for all models on an example tile from PASTIS-24.}
\label{fig:appBpastis24b2104}
\end{figure*}

\FloatBarrier

\onecolumn

\section{Confusion Matrices of PAtteRNS, TSViT, \& UTAE}\label{app:confmatrices}

\begin{figure*}[htb]
\centering
\includegraphics[width=0.84\textwidth]{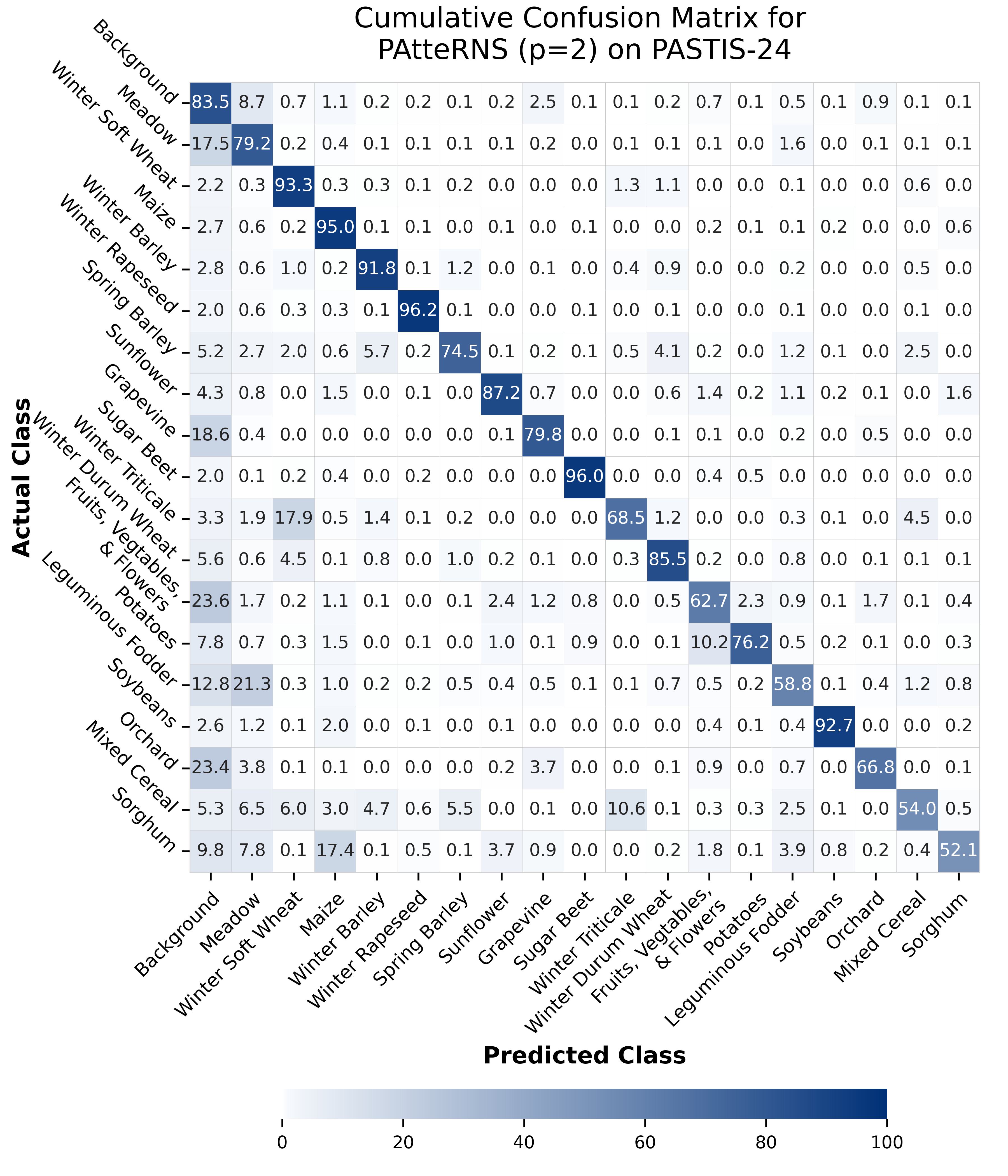}
\caption{The cumulative class-wise confusion matrix of all seeded repeats and folds for PAtteRNS $p=2$ on PASTIS 24}
\label{fig:appCconfmatrixPASTIS24PAtteRNSP2}
\end{figure*}

\begin{figure*}[htb]
\centering
\includegraphics[width=0.84\textwidth]{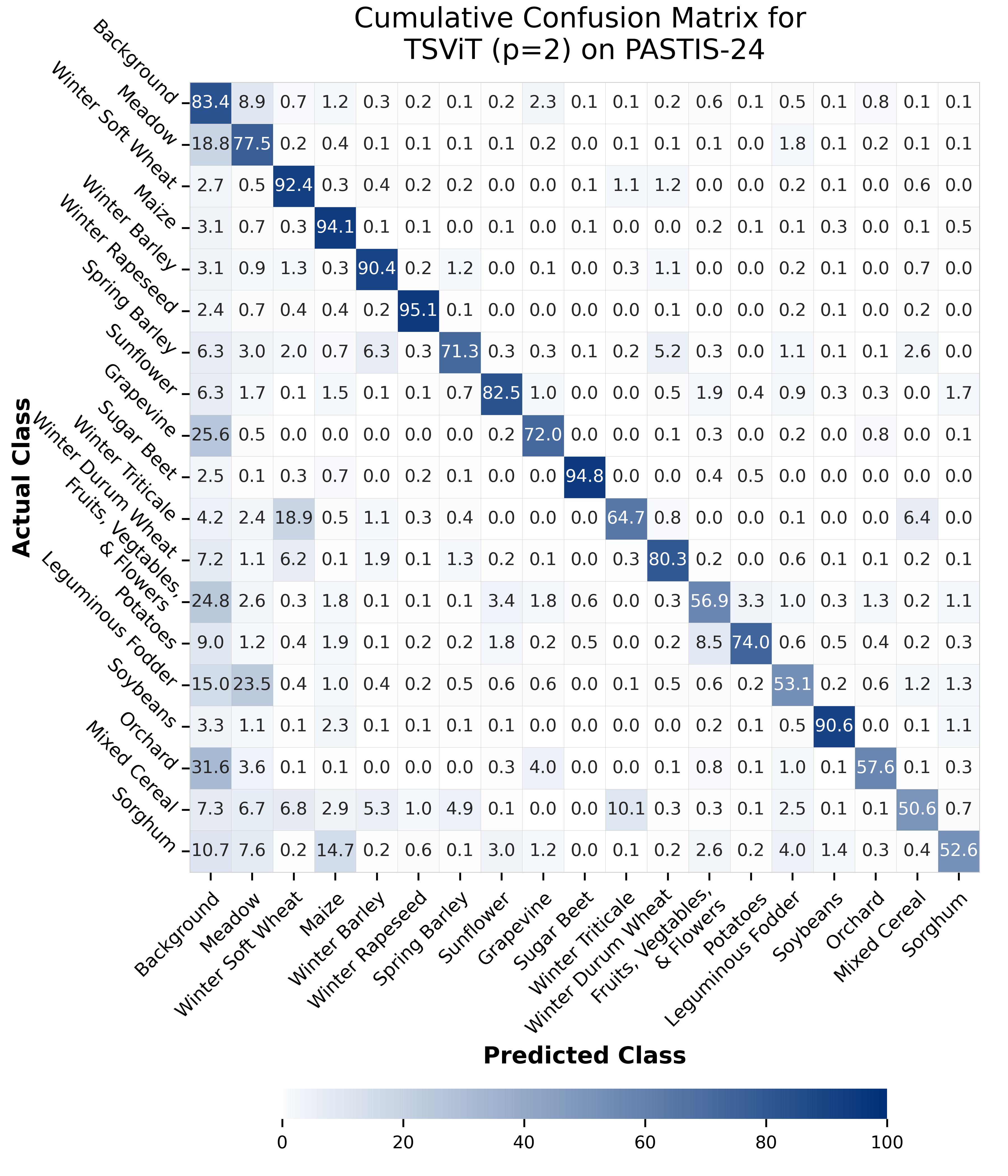}
\caption{The cumulative class-wise confusion matrix of all seeded repeats and folds for TSViT $p=2$ on PASTIS 24}
\label{fig:appCconfmatrixPASTIS24TSViTP2}
\end{figure*}

\begin{figure*}[htb]
\centering
\includegraphics[width=0.84\textwidth]{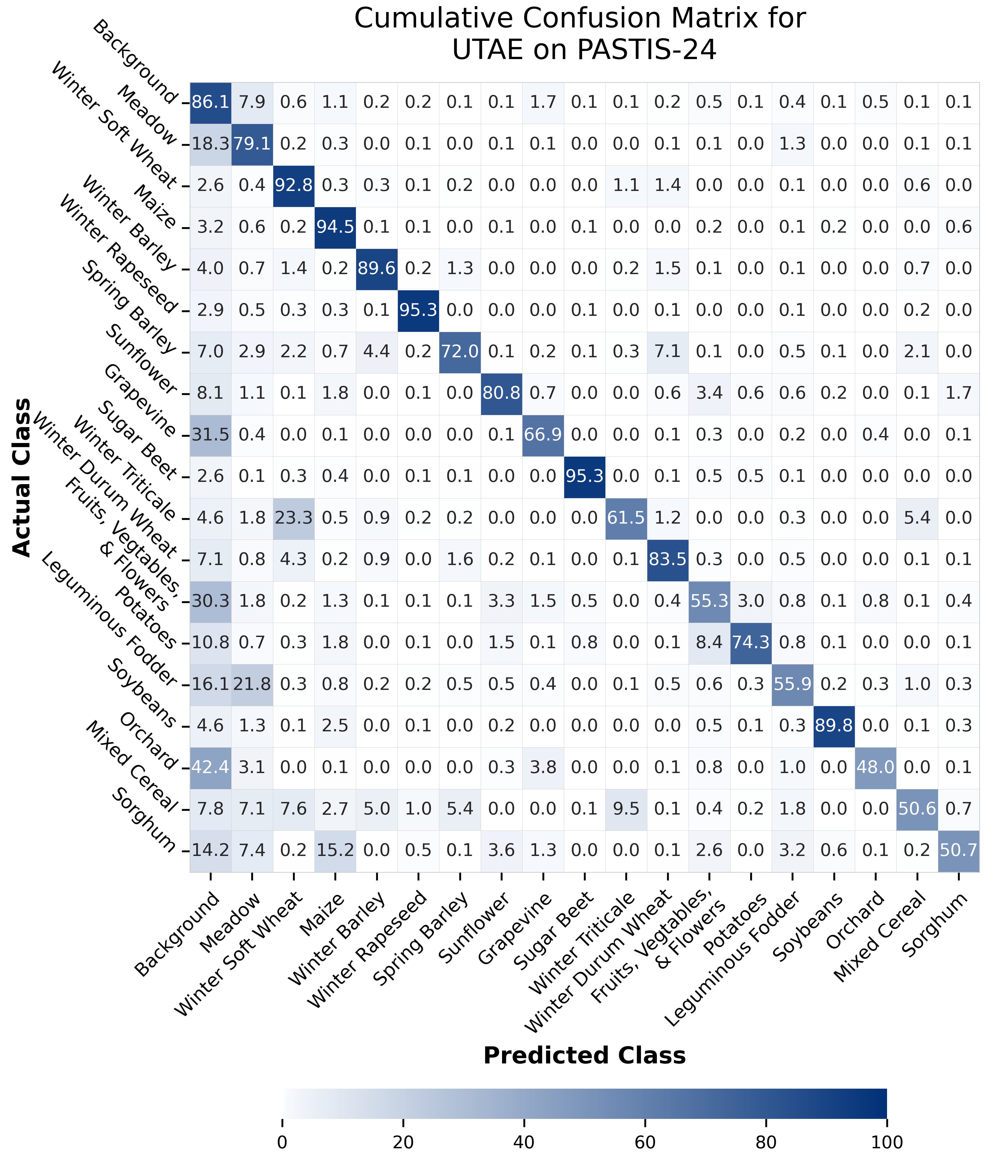}
\caption{The cumulative class-wise confusion matrix of all seeded repeats and folds for UTAE on PASTIS 24}
\label{fig:appCconfmatrixPASTIS24UTAE}
\end{figure*}

\begin{figure*}[htb]
\centering
\includegraphics[width=0.84\textwidth]{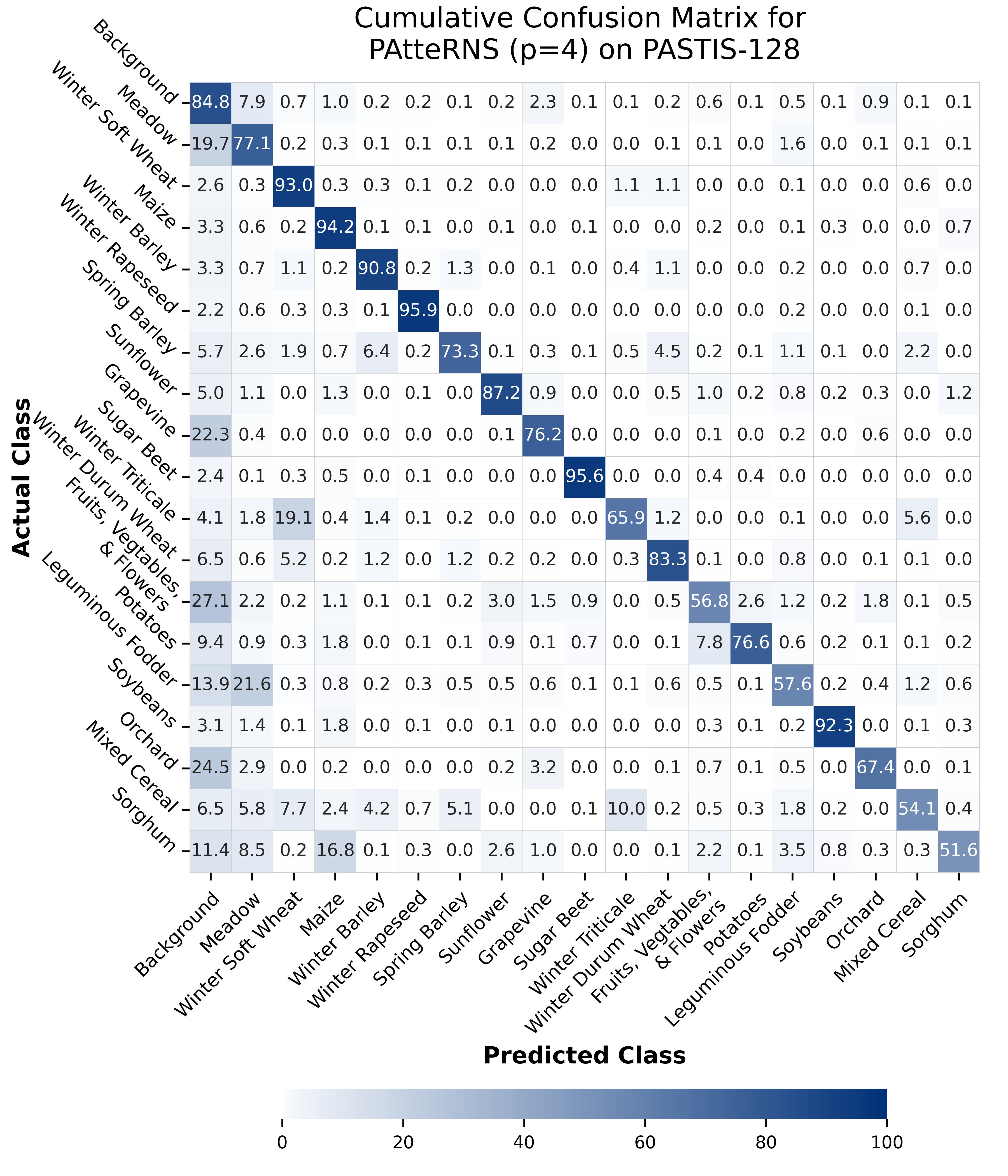}
\caption{The cumulative class-wise confusion matrix of all seeded repeats and folds for PAtteRNS $p=4$ on PASTIS 128}
\label{fig:appCconfmatrixPASTIS128PAtteRNSP4}
\end{figure*}

\begin{figure*}[htb]
\centering
\includegraphics[width=0.84\textwidth]{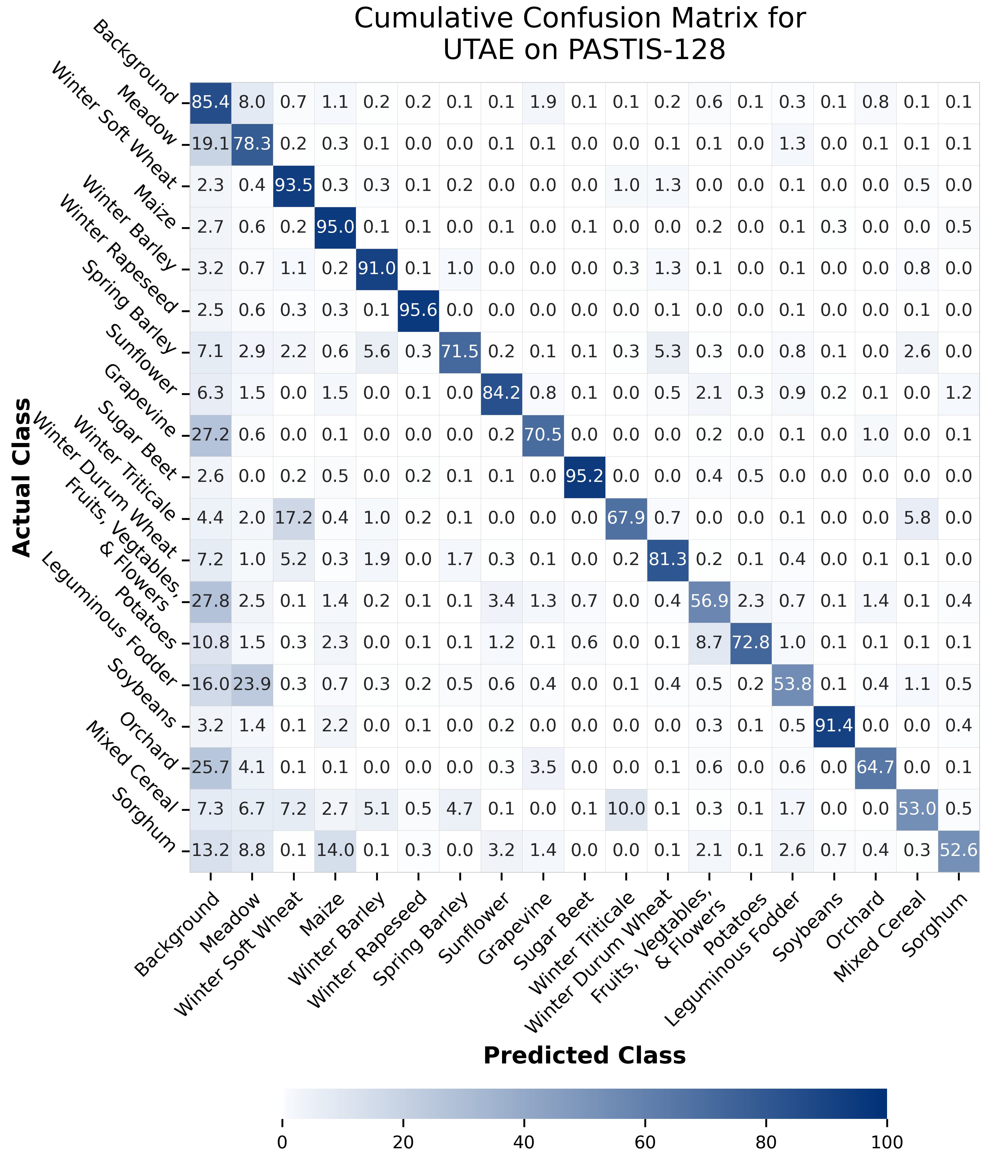}
\caption{The cumulative class-wise confusion matrix of all seeded repeats and folds for UTAE on PASTIS 128}
\label{fig:appCconfmatrixPASTIS128UTAE}
\end{figure*}

\begin{figure*}[htb]
\centering
\includegraphics[width=0.84\textwidth]{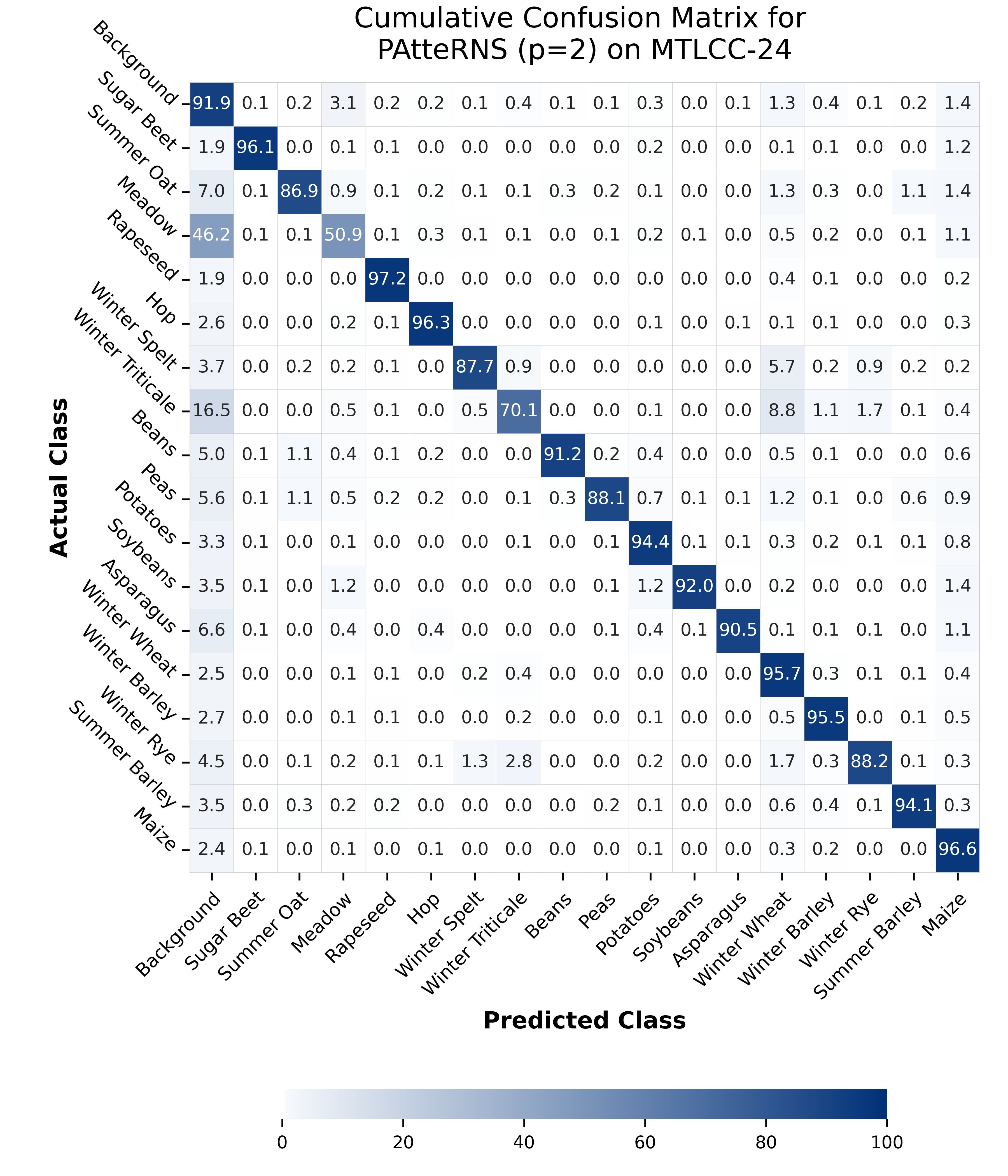}
\caption{The cumulative class-wise confusion matrix of all seeded repeats and folds for PAtteRNS $p=2$ on MTLCC 24}
\label{fig:appCconfmatrixMTLCC24PAtteRNSP2}
\end{figure*}

\begin{figure*}[htb]
\centering
\includegraphics[width=0.84\textwidth]{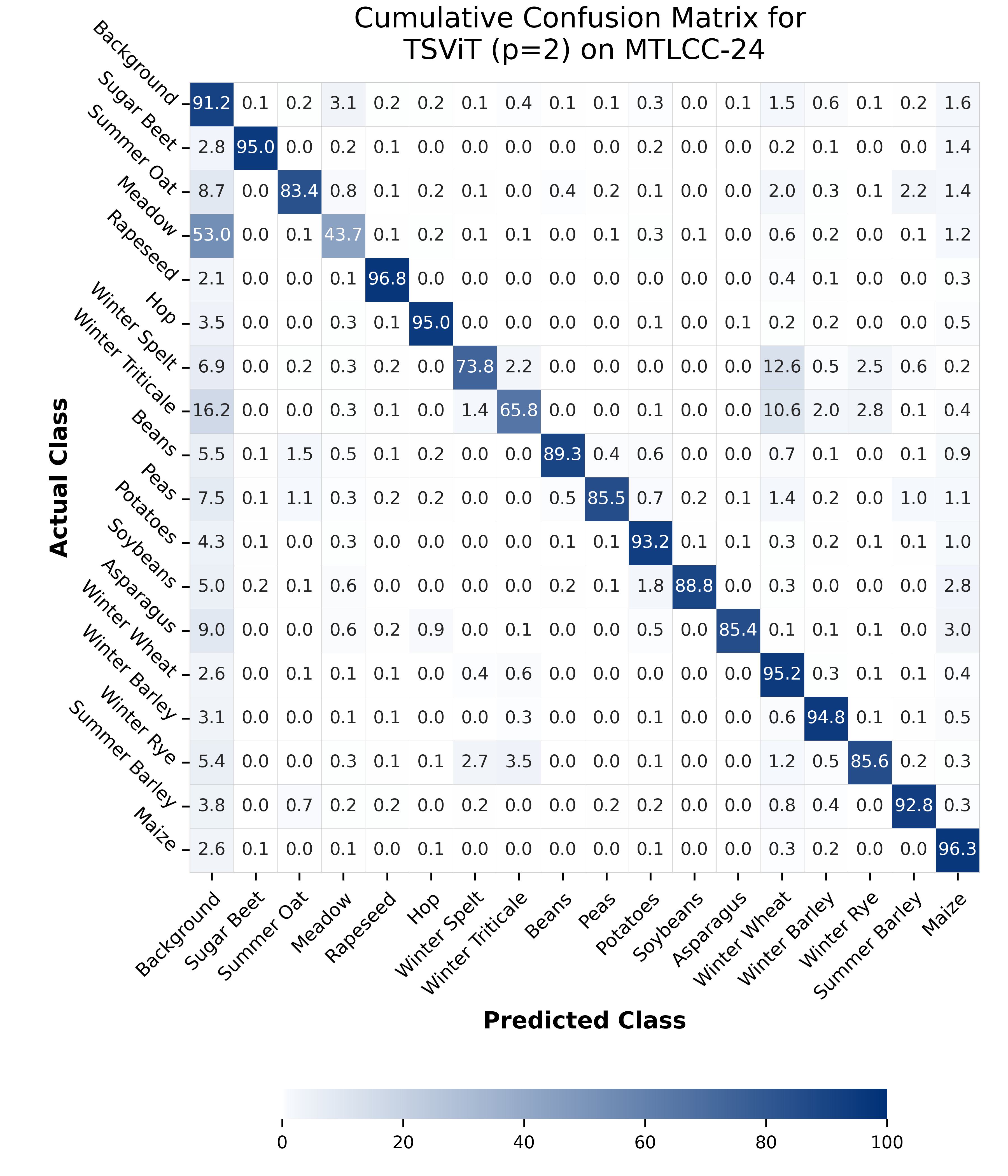}
\caption{The cumulative class-wise confusion matrix of all seeded repeats and folds for TSViT $p=2$ on MTLCC 24}
\label{fig:appCconfmatrixMTLCC24TSViTP2}
\end{figure*}

\begin{figure*}[htb]
\centering
\includegraphics[width=0.84\textwidth]{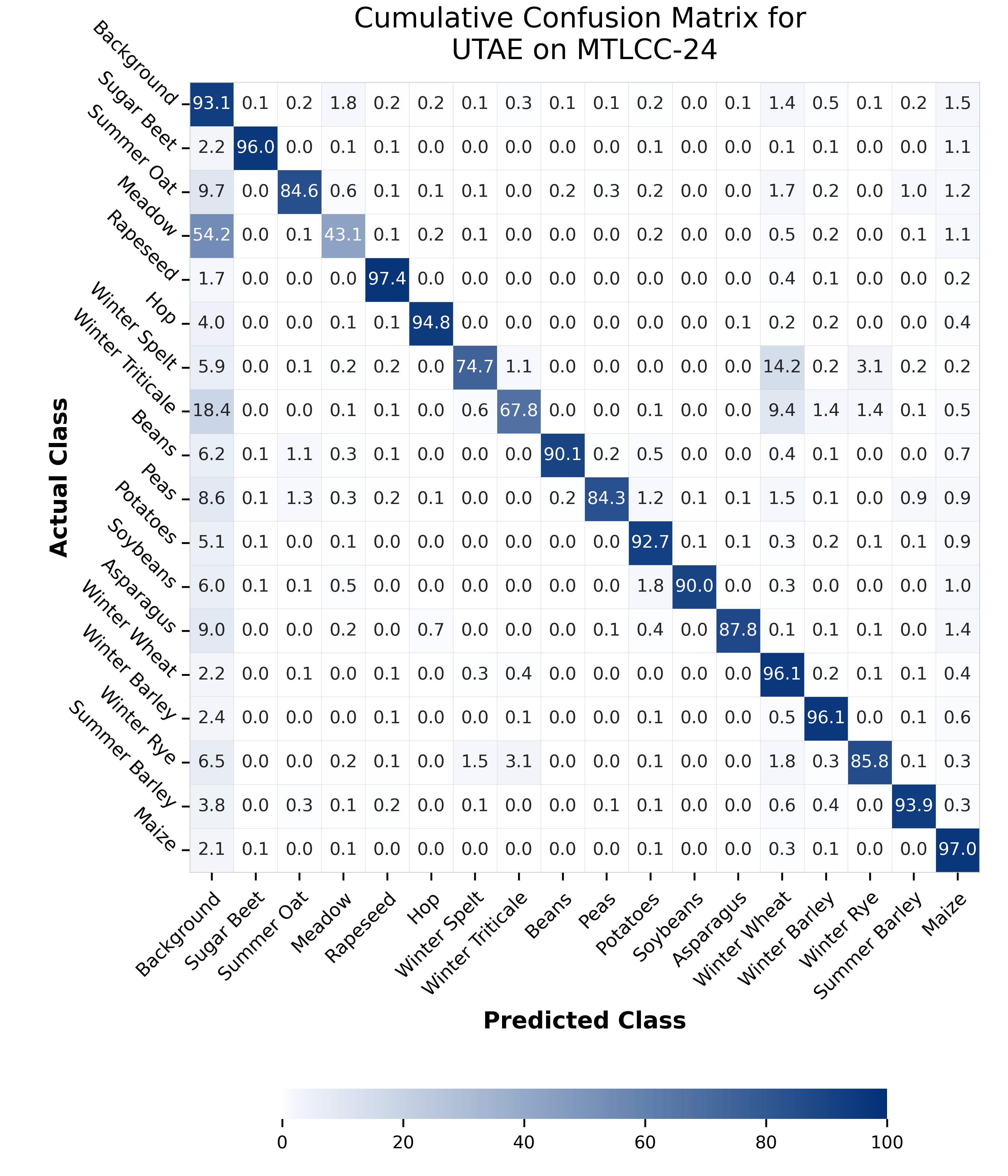}
\caption{The cumulative class-wise confusion matrix of all seeded repeats and folds for UTAE on MTLCC 24}
\label{fig:appCconfmatrixMTLCC24UTAE}
\end{figure*}

\end{document}